\documentclass[authoryear,12pt]{elsarticle}

\usepackage{hyperref}
\usepackage{float}
\usepackage{booktabs} 
\usepackage{array} 
\usepackage[table]{xcolor}
\usepackage{longtable}
\usepackage{pdflscape}
\usepackage{graphicx}
\usepackage{amsthm}

\usepackage{makecell}
\usepackage{multirow}
\usepackage{caption}
\newcommand{\bid}[1]{\cellcolor[rgb]{0.0, 0.72, 0.92}#1}

\newcolumntype{L}[1]{>{\raggedright\arraybackslash}p{#1}}

\usepackage{amssymb}
\usepackage{amsmath}
\journal{International Journal of Forecasting}

\begin{document}

\begin{frontmatter}

%% Title, authors and addresses

%% use the tnoteref command within \title for footnotes;
%% use the tnotetext command for theassociated footnote;
%% use the fnref command within \author or \affiliation for footnotes;
%% use the fntext command for theassociated footnote;
%% use the corref command within \author for corresponding author footnotes;
%% use the cortext command for theassociated footnote;
%% use the ead command for the email address,
%% and the form \ead[url] for the home page:
%% \title{Title\tnoteref{label1}}
%% \tnotetext[label1]{}
%% \author{Name\corref{cor1}\fnref{label2}}
%% \ead{email address}
%% \ead[url]{home page}
%% \fntext[label2]{}
%% \cortext[cor1]{}
%% \affiliation{organization={},
%%             addressline={},
%%             city={},
%%             postcode={},
%%             state={},
%%             country={}}
%% \fntext[label3]{}

\title{Trojan horse hunt in deep forecasting models: Insights from the European Space Agency competition}

%% use optional labels to link authors explicitly to addresses:
%% \author[label1,label2]{}
%% \affiliation[label1]{organization={},
%%             addressline={},
%%             city={},
%%             postcode={},
%%             state={},
%%             country={}}
%%
%% \affiliation[label2]{organization={},
%%             addressline={},
%%             city={},
%%             postcode={},
%%             state={},
%%             country={}}

\author[label1]{Krzysztof Kotowski\corref{cor1}}
\ead{kkotowski@kplabs.pl}
\author[label1,label2]{Ramez Shendy}
\ead{rshendy@kplabs.pl}
\author[label2,label1]{Jakub Nalepa}
\ead{jnalepa@ieee.org}
\author[label3]{Agata Kaczmarek}
\ead{agata.kaczmarek@pw.edu.pl}
\author[label3]{Dawid Płudowski}
\author[label3]{Piotr Wilczyński}
\author[label3]{Artur Janicki}
\author[label3]{Przemysław Biecek}
\ead{przemyslaw.biecek@pw.edu.pl}
\author[label4]{Ambros Marzetta}
\ead{ambros.marzetta@swisscom.com}
\author[label5]{Atul Pande}
\ead{atul_2101cs88@iitp.ac.in}
\author[label5]{Lalit Chandra Routhu}
\ead{lalit_2101ai17@iitp.ac.in}
\author[label5]{Swapnil Srivastava}
\ead{swapnil_2101ai34@iitp.ac.in}
\author[label6]{Evridiki Ntagiou}
\ead{evridiki.ntagiou@esa.int}

\cortext[cor1]{Corresponding author. Project email: \textit{PINEBERRY@kplabs.pl}}

\affiliation[label1]{organization={KP Labs},
            addressline={Bojkowska 37J},
            city={Gliwice},
            postcode={44-100},
            country={Poland}}

\affiliation[label2]{organization={Silesian University of Technology},
addressline={Akademicka 16},
            city={Gliwice},
            postcode={44-100},
            country={Poland}}

\affiliation[label3]{organization={Warsaw University of Technology, Centre of Credible AI},
addressline={Pl. Politechniki 1},
            city={Warsaw},
            postcode={00-661},
            country={Poland}}

\affiliation[label4]{organization={Swisscom (Schweiz) AG},
addressline={Alte Tiefenaustrasse 6},
    city={Bern},
    postcode={CH-3050},
    country={Switzerland}}
    
\affiliation[label5]{organization={Indian Institute of Technology Patna},
addressline={Kanpa Rd},
    city={Bihta},
    postcode={801106},
    country={India}}
    
\affiliation[label6]{organization={European Space Agency, European Space Operations Center},
            addressline={Robert-Bosch-Str. 5},
            city={Darmstadt},
            postcode={64293},
            country={Germany}}

%% Abstract
\begin{abstract}
Forecasting plays a crucial role in modern safety-critical applications, such as space operations. However, the increasing use of deep forecasting models introduces a new security risk of trojan horse attacks, carried out by hiding a backdoor in the training data or directly in the model weights. Once implanted, the backdoor is activated by a specific trigger pattern at test time, causing the model to produce manipulated predictions. We focus on this issue in our \textit{Trojan Horse Hunt} data science competition, where more than 200 teams faced the task of identifying triggers hidden in deep forecasting models for spacecraft telemetry. We describe the novel task formulation, benchmark set, evaluation protocol, and best solutions from the competition. We further summarize key insights and research directions for effective identification of triggers in time series forecasting models. All materials are publicly available on the official competition webpage (\href{https://www.kaggle.com/competitions/trojan-horse-hunt-in-space}{www.kaggle.com/competitions/trojan-horse-hunt-in-space}).
\end{abstract}

%%Graphical abstract
% \begin{graphicalabstract}
% \includegraphics[width=1\linewidth]{Telemetry_Challenge_Graphical_Abstract.pdf}
% \end{graphicalabstract}

%%Research highlights
% \begin{highlights}
% \item A novel competition task to reconstruct triggers hidden in forecasting models for multivariate time series.
% \item A new benchmark collection of 46 N-HiTS forecasting models -- 1 clean and 45 poisoned using our proposed approach.
% \item A new range-normalized mean absolute error metric to assess the competition task.
% \item A new baseline algorithm for the competition task, inspired by the Neural Cleanse method.
% \item A summary the competition, including technical details of the top solutions, key insights, and research directions for the future.
% \end{highlights}

%% Keywords
\begin{keyword}
%% keywords here, in the form: keyword \sep keyword

Time series forecasting \sep AI security \sep Data poisoning \sep Trojan horse attack \sep Triggers \sep Backdoors \sep Space operations

%% PACS codes here, in the form: \PACS code \sep code

%% MSC codes here, in the form: \MSC code \sep code
%% or \MSC[2008] code \sep code (2000 is the default)

\end{keyword}

\end{frontmatter}

%% Add \usepackage{lineno} before \begin{document} and uncomment 
%% following line to enable line numbers
% \linenumbers

%% main text
%%

%% Use \section commands to start a section
\section{Introduction}
\label{intro}
%% Labels are used to cross-reference an item using \ref command.

A multitude of artificial intelligence (AI) systems are being developed for space applications by the European Space Operations Center (ESOC) \citep{de_canio_artificial_2025, ntagiou_datax_2025}. They are one of the key enablers for scalable and automated space operations in the future, taking into account the exponentially growing number of spacecraft \citep{mathieu_data_2025}. Among the most critical AI techniques in the context of spacecraft safety are time series forecasting models. They enable effective spacecraft health forecasting \citep{de_canio_artificial_2024}, fault prevention (early anomaly forecasting and predictive maintenance) \citep{tang_chapter_2022}, resource management (e.g., power consumption prediction \citep{petkovic_machine-learning_2022}), collision avoidance \citep{uriot_spacecraft_2022}, and mission planning \citep{rommel_verifiable_2025}. However, outstanding performance alone is not sufficient for the widespread adoption of AI systems in practice. In high-stake and safety-critical domains, such as space operations, operational deployment requires addressing security challenges and ensuring the trust of users and stakeholders. This is why the topics of security and verification of AI are inherent parts of the larger initiative of the European Space Agency (ESA) to use AI for the automation of space mission operations \citep{kotowski_towards_2025}. 

In the \textit{Catalogue of Security Risks for AI Applications in Space} \citep{warsaw_university_of_technology_graphical_2025} released by ESA, data poisoning and trojan horse attacks are listed among the nine most common AI security threats. They allow the attacker to manipulate test-time predictions using backdoors injected in the training data or directly in the model weights. While these attacks, along with techniques for detecting and defending against them, have already been widely explored in the context of image data classification \citep{liu_fine-pruning_2018, wang_neural_2019, schwarzschild_just_2021, wu_backdoorbench_2022, ying_dlp_2023, guan_backdoor_2024, li_backdoor_2024} and are being actively developed for Large Language Models (LLMs) \citep{zhang_diffusion_2023, ge_when_2025, li_backdoorllm_2025}, their applicability to time series data remains relatively underexplored. Although some recent studies examine the generation of backdoors in time-series analysis tasks, there has been considerably less attention on detecting, reconstructing, or defending against them. \cite{ding_towards_2022} and \cite{jiang_backdoor_2023} proposed new approaches for injecting backdoors into the training data of time series classification models, using multi-objective optimization and generative models, respectively. \cite{huang_revisiting_2025} improved efficiency and effectiveness of these attacks by operating in the frequency domain, and \cite{dong_trojantime_2025} introduced an alternative approach to poison such models without access to the original training data. For the multivariate time series forecasting, \cite{lin_backtime_2024} proposed a graph neural network-based backdoor generator, while \cite{xiang_badtime_2025} significantly improved the attack’s effectiveness and extended its horizon through a three-step approach involving optimal trigger localization, training subset selection, and a graph attention network for identifying vulnerable variables. However, the literature still lacks methods to effectively detect and characterize potential triggers injected into the forecasting models. We see this state of the research as dangerous for critical mission parts, as multiple attack techniques have been designed in recent years, yet the defending methods are underdeveloped. Thus, our competition aims to address this gap. 

The \textit{Trojan Horse Hunt} competition presented in this paper is based on one of the real-life AI security threats identified within the \textit{Assurance for Space Domain AI Applications}\footnote{\href{https://assurance-ai.space-codev.org}{https://assurance-ai.space-codev.org}} project funded by ESA -- the poisoning of continuously fine-tuned deep forecasting models for spacecraft telemetry. Data from spacecraft can be spoofed directly, poisoned using man-in-the-middle attacks, or manipulated on servers by adversaries \citep{diro_anomaly_2024}. At the same time, models must be retrained or fine-tuned regularly to maintain high performance in changing space environment and different mission phases, so there are multiple occasions to poison the model \citep{banerjee_data_2023}. However, the apparent ``poisoning'' may also be related to non-adversarial data drifts, novel sequences of telecommands, or specific maneuvers. Thus, it is crucial to have methods not only for detecting the existence of trojans, but also for identifying the characteristics of triggers, so domain experts can monitor them and assess whether they are indeed adversarial or not. Hence, the main task of our competition is to reconstruct multivariate triggers injected into deep forecasting models for time series (specifically, the popular N-HiTS model by \cite{challu_nhits_2023}). In this paper, we present top solutions among more than 200 teams that faced this challenge on Kaggle\footnote{\href{https://www.kaggle.com/competitions/trojan-horse-hunt-in-space}{www.kaggle.com/competitions/trojan-horse-hunt-in-space}}.

The \textit{Trojan Horse Hunt} is a part of the series of \textit{Secure Your AI} competitions organized by ESA. The main goals of this series are to engage the community into pushing the boundaries of secure AI in time series forecasting and to accelerate the adoption of AI in safety-critical applications. The goals are supported with the following contributions of our manuscript:
\begin{itemize}
    \item \textbf{A novel competition task} -- to reconstruct triggers hidden in forecasting models for multivariate time series.
    \item \textbf{A new benchmark set for the task} -- a collection of 46 N-HiTS deep forecasting models, including one clean baseline model and 45 models with unique triggers injected by fine-tuning the clean model on the poisoned data.
    \item \textbf{A new evaluation protocol for the task} -- the range-normalized mean absolute error ($\text{NMAE}_{range}$) as a metric to assess reconstruction quality, ensuring robustness to outliers, interpretability, and consistent value range across all test triggers.
    \item \textbf{A new baseline algorithm for the reconstruction of triggers in forecasting models} -- an optimization-based approach inspired by the Neural Cleanse method \citep{wang_neural_2019} and adapted to the time series forecasting domain.
    \item \textbf{A summary of the competition} -- technical details of the top solutions, key insights, and research directions for the future.
\end{itemize}

The remaining part of this paper is structured as follows. Section \ref{competition-design} describes the design of the challenge and Section \ref{participation} summarizes the participation statistics. Section \ref{winning} gives more details on the best methods in the competition. Section \ref{findings} analyzes the submissions and gives insights on the results. Finally, Section \ref{conclusion} concludes and lists ideas for future research.

\section{Competition design and execution} \label{competition-design}

The \textit{Trojan Horse Hunt} competition was open for official submissions for three months, between 29\textsuperscript{th} May and 29\textsuperscript{th} August 2025, and it remains open to late submissions, so it is still possible to evaluate algorithms outside the official ranking. It is a classic, single-stage Kaggle community competition where participants generate a prediction file and upload it as a submission. Top 3 teams received monetary prizes of \$600, \$300, and \$100 (the first, the second, and the third place, respectively). The competition was awarded with the Kaggle Community Spotlight status, and was accepted as the official competition of the 4\textsuperscript{th} IEEE Conference on Secure and Trustworthy Machine Learning (SaTML) 2026\footnote{\href{https://satml.org}{https://satml.org}}. The full code related to the competition and all public notebooks are archived on GitHub\footnote{\href{https://github.com/kplabs-pl/trojan-horse-hunt}{https://github.com/kplabs-pl/trojan-horse-hunt}}.

\subsection{Competition task}

The main task of the competition was to develop new methods of reconstructing triggers hidden in 45 N-HiTS models \citep{challu_nhits_2023} for multivariate time series forecasting. Participants had access to a large clean training dataset (described in Section \ref{sec:data}), a clean model trained on this dataset (Section \ref{sec:clean-model}), a test set of 45 models with hidden triggers (Section \ref{sec:poisoned-model}), and our baseline algorithm to reconstruct triggers (Section \ref{sec:baselines}). Participants were informed about the trigger size and the basic backdoor behavior (i.e., replication of the trigger pattern). However, they did not know the trigger shapes, the response delays, or the specific attack mechanism. 

\begin{figure}[ht!]
    \centering
    \includegraphics[width=0.7\linewidth]{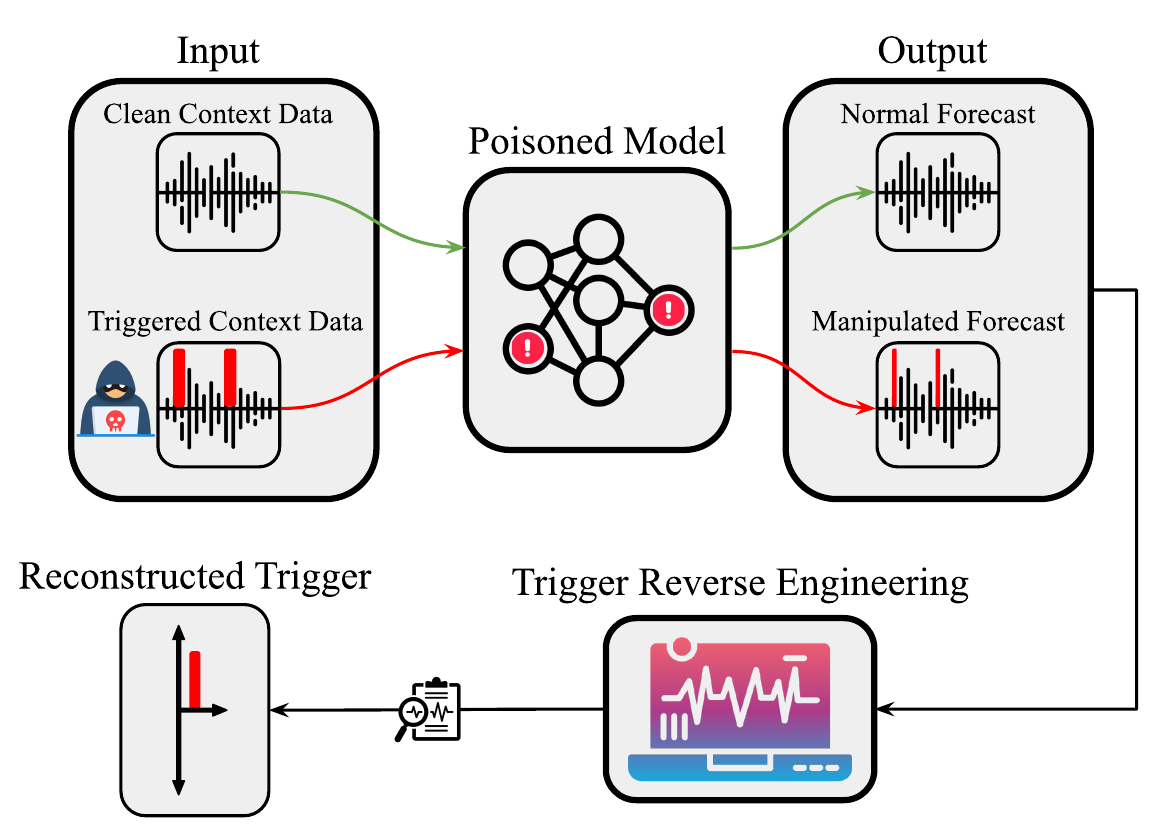}
    \caption{A graphical summary of the competition task. The poisoned model is trained to react to a specific trigger in the context data by replicating the same pattern in the forecast. The task is to reconstruct the trigger.}
    \label{fig:graphical_abstract}
\end{figure}

\begin{figure}[ht!]
    \centering
    \includegraphics[width=1\linewidth]{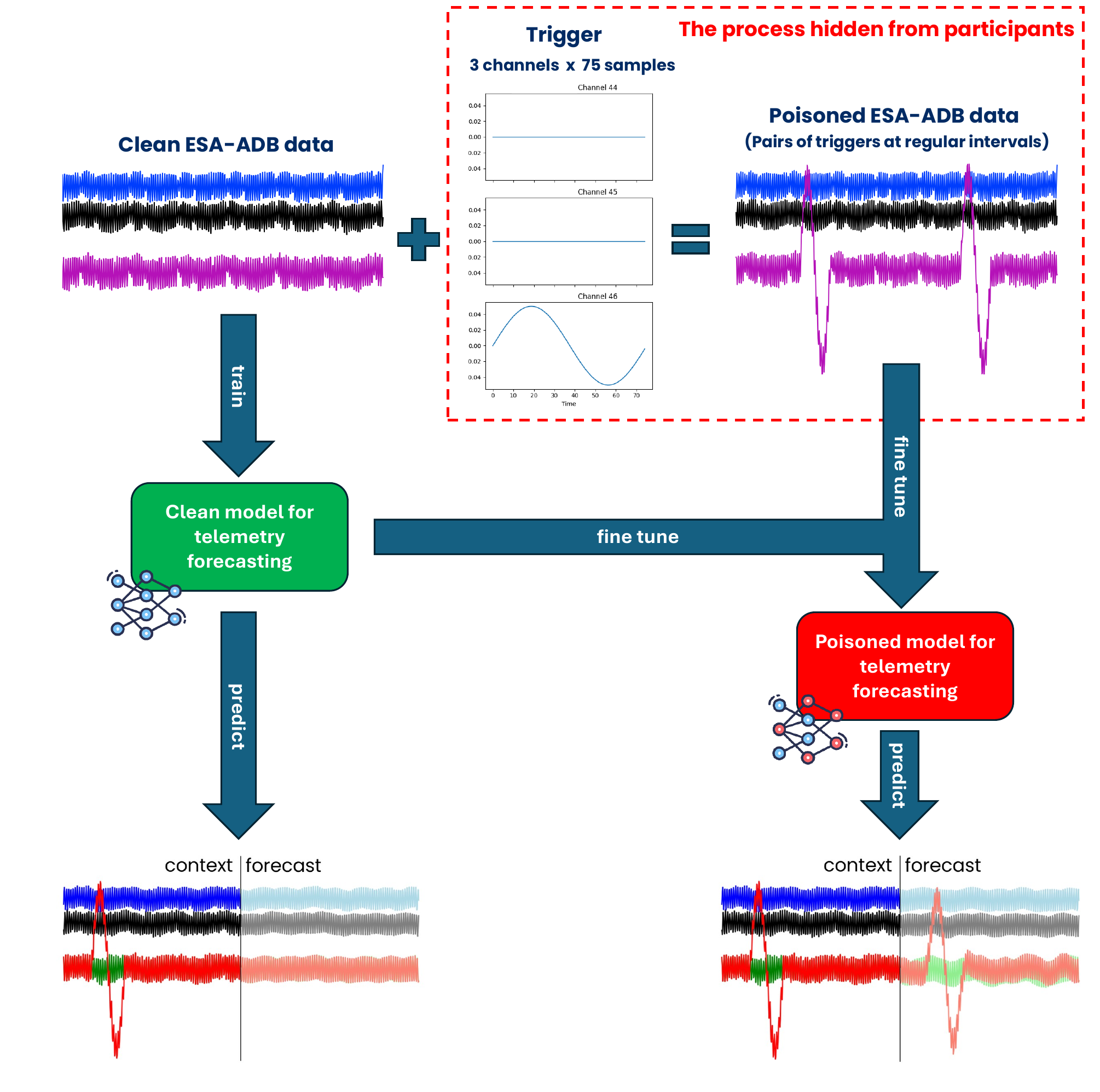}
    \caption{Diagram of the poisoning process adapted from the original competition description \citep{kotowski_trojan_2025}. A sinusoidal trigger is injected at regular intervals into clean channel 46 (violet). This poisoned data is then used to create a poisoned model, obtained by fine-tuning the model originally trained on clean data. Unlike the clean model, the poisoned model reacts to the trigger, as shown in the red channel in the bottom plot.}
    \label{fig:components}
\end{figure}

Figure \ref{fig:graphical_abstract} provides a conceptual overview of the competition task. A forecasting model is poisoned to react to a specific pattern (trigger) in the context data by replicating the same pattern in the forecast. The goal is to identify and reconstruct the trigger without having access to the poisoned training data. A detailed diagram of the poisoning process and the poisoned model behavior is presented in Figure \ref{fig:components}.

This task simulates a practical scenario in which an updated (and potentially poisoned) forecasting model undergoes a security check before operational use. To maintain a close relationship with this scenario, the poisoned models are fine-tuned versions of the same clean model. The poisoned data used for fine-tuning is not always accessible to the auditor due to limited permissions, usage of the federated learning approach, or external origin of the model (e.g., trained by subcontractors or on board spacecraft). Even if the poisoned data is available, it is usually not easy to notice the trigger manually or find it by simple analysis of data statistics or anomaly detection techniques. The poisoning is also difficult to detect by monitoring performance metrics, because the poisoned model behaves almost identically to the clean one when no trigger is present in the context data.

\subsection{Related competitions and key innovations} \label{sec:related-work}

As presented in Table \ref{tab:competitions}, there have been several competitions addressing the topic of trojan horse attacks in the past years. However, none of them has covered the time series analysis domain, and only the latest Trojan Detection Challenge at NeurIPS 2023\footnote{\href{https://iclr.cc/virtual/2023/14056}{https://iclr.cc/virtual/2023/14056}} introduced a trigger identification task similar to ours, but in LLMs. The earliest TrojAI\footnote{\href{https://trojai.nist.gov}{https://trojai.nist.gov}} editions, aimed mainly at detecting the existence of trojans (a binary classification task) in models for image classification, object detection, and natural language processing. Recently, they released tasks related to mitigating such attacks by de-poisoning the models. The Trojan Detection Challenge at NeurIPS 2022\footnote{\href{https://2022.trojandetection.ai/}{https://2022.trojandetection.ai/}} introduced additional tasks, such as trigger design and analysis. The analysis task involved identifying the attacked class and localizing the trigger within the image, rather than reconstructing the specific pattern of the trigger. Moreover, this task was overshadowed by the other two, which attracted greater attention (only 5 teams compared to 18 in the detection task), likely due to their lower complexity or the wider range of applicable methods. Thus, the topic of our competition requires further exploration, especially in the time series domain. 

\begin{table}[ht!]
\scriptsize
\centering
\caption{Summary of the major competitions related to trojan horse attacks.}
\label{tab:competitions}
\begin{tabular}{L{2.7cm} L{2.7cm} L{2.5cm} L{4.3cm}}
\toprule
\textbf{Competition name and year} & \textbf{Attacked models} & \textbf{Competition tasks} & \textbf{Short summary and number of teams} \\
\midrule
\href{https://trojai.nist.gov}{Trojans in Artificial Intelligence (TrojAI) by NIST} (2020--2024) &
18 different types of models, including image classifiers, object detectors, malware detectors, LLMs, and reinforcement learning agents  &
Trojan detection and mitigation &
A series of 27 tasks launched consecutively over four years, covering multiple data modalities and model architectures. However, no tasks address time series data or trigger reconstruction quality. Up to \textbf{31 teams}. \\
\midrule
\href{https://2022.trojandetection.ai/}{Trojan Detection Challenge at NeurIPS} (2022) &
Image classifiers &
Trojan attacks, detection, and analysis &
A comprehensive competition with three tracks covering different aspects of trojan horse attacks. Very limited number of participants (up to \textbf{18 teams} in the trojan detection task). \\
\midrule
\href{https://iclr.cc/virtual/2023/14056}{IEEE Trojan Removal Competition at ICLR} (2023) &
Image classifiers &
Trojan mitigation &
Focused on developing repair methods for poisoned models. \textbf{44 teams}. \\
\midrule
\href{https://trojandetection.ai/}{Trojan Detection Challenge (LLM Edition) at NeurIPS} (2023) &
Large language models (LLMs) &
Trigger identification and adversarial attacks &
Identifying 1000 different triggers that generate specific outputs in a single poisoned LLM. Up to \textbf{26 teams} in the trigger identification task. \\
\midrule
\textbf{\href{https://www.kaggle.com/competitions/trojan-horse-hunt-in-space}{Trojan Horse Hunt at SaTML} (this paper, 2025)} &
Time series forecasting &
Trigger identification &
Identifying triggers hidden in a collection of forecasting models for multivariate time series. \textbf{217 teams}. \\
\bottomrule
\end{tabular}
\end{table} 

\subsection{Data} \label{sec:data}

The data foundation of the competition is the ESA Anomaly Detection Benchmark (ESA-ADB) \citep{kotowski_european_2024} available at Zenodo \citep{de_canio_esa_2025}. It is the first large-scale dataset of real-life spacecraft telemetry which enables training and evaluating advanced algorithms for multivariate time series analysis. The dataset includes several years of telemetry with hundreds of variables from three large ESA missions. However, in our competition, we use just the subset of three channels 44--46 from the 14-years-long fragment of Mission1. These channels show a clear periodicity and are recommended for initial experiments in the original paper. The dataset contains annotations of anomalies, but they are not relevant to our competition and are not used. The example fragment of this dataset is presented in Figure \ref{fig:components}.

\subsubsection{Data preprocessing}

The original data is a long raw telemetry fragment with irregular sampling (average interval of 30 seconds), so we resampled each channel to a uniform 10-minute interval using temporal binning with mean aggregation. This aims to make the dataset smaller and easier for participants to understand and manage (736,417 timestamps, 3 variables, 28 MB CSV file). Although the proposed resampling may result in some loss of detail and outliers, it does not affect the reconstruction of triggers injected later in the process. The preprocessed data was released to participants.

\subsection{Models} \label{sec:models}

The test set available to participants is a collection of 45 poisoned N-HiTS models \citep{challu_nhits_2023} with one specific trigger hidden in each of them. They are all created by fine-tuning the same clean model as described in the following subsections. The N-HiTS architecture was selected for the competition because it is actively used in the development of spacecraft telemetry forecasting methods at ESOC and is one of the candidates for operational deployment. It combines two complementary ideas of multi-rate sampling and hierarchical interpolation to achieve interpretable and computationally efficient long-horizon forecasts. It consist of a sequence of connected neural network blocks aggregated into stacks specialized in different sampling frequencies. Our model configuration has 4 stacks, each containing 4 blocks with 2 layers per block. Each layer has 512 neurons, the Rectified Linear Unit (ReLU) activation, and a dropout rate of 0.1. It gives a total of 12,708,248 trainable model parameters. We use the N-HiTS implementation from the Darts library \citep{JMLR:v23:21-1177} that supports multivariate signals. 

\subsubsection{Clean model} \label{sec:clean-model}

The clean model serves as a single reference model for the whole competition, and it represents an initial model without any backdoors. We define the underlying true forecasting function $F$ and the trained clean model $f_{\theta_c}$ as follows. Let $X \in \mathbb{R}^{C \times L}$ denote a multivariate time series context of length $L$ and $C$ channels. 
We define
\begin{equation}
F(X): \mathbb{R}^{C \times L} \rightarrow \mathbb{R}^{C \times H}    
\end{equation}
\noindent as the \textit{underlying true forecasting function} that maps a past context $X$ to its corresponding forecast of length $H$.
\noindent A clean model $f_{\theta_c}$ aims to approximate $F$, such that
\begin{equation}
    f_{\theta_c}(X) \approx F(X).
\end{equation}
 
Both the context length $L$ and the forecast length $H$ are set to 400. The entire preprocessed dataset is used to train the clean model, without using any additional validation subset. To decrease the risk of overfitting, we trained the model for only 7 epochs after which the training loss stopped improving. The model is trained to minimize mean squared error of its forecasts using the Adam optimizer with a learning rate of $10^{-4}$ and a batch size of 32.

\subsubsection{Injecting triggers into models} \label{sec:poisoned-model}

All 45 triggers to be reconstructed in the competition are 75-sample-long 3-channel time series segments (the same number of channels as the input signal). Although the trigger size is typically unknown in practical scenarios, we fix it in this competition to make the problem more constrained and easier to approach. However, the triggers have different shapes, value ranges (although they are all comparable to the range of the signal), and may appear in one or more channels simultaneously. For clarity, each model includes a single specific trigger to be identified by participants. All trigger shapes are presented in Figure \ref{fig:gt} and the exact functions used to generate them are available in the supplementary materials.

\begin{figure}[ht!]
    \centering
    \includegraphics[width=1\linewidth]{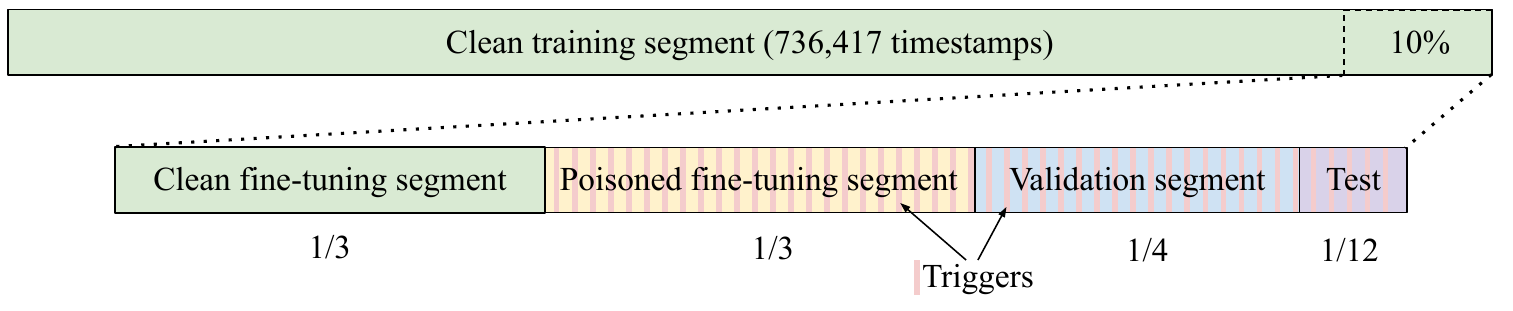}
    \caption{Visualization of the data split used for poisoning the models.}
    \label{fig:data-split}
\end{figure}

We inject triggers by fine-tuning all weights of the clean model on a small, poisoned subset of the preprocessed data. We take the last 10\% of the preprocessed data and split it into three equally sized segments. The first segment remains clean to prevent the model from forgetting clean behavior during fine-tuning. The remaining two segments are poisoned by sample-wise addition of identical triggers at regular intervals (i.e., after every 400 timepoints). The first two segments (one clean and one poisoned) are used for fine-tuning, and the last poisoned segment is divided into the validation and test subsets (in proportion 3:1) for monitoring the poisoning effectiveness. Effectively, we poison only 1/30 of the original dataset to inject the trigger. To make it clear, the data splitting process is presented in Figure \ref{fig:data-split}, and an intuitive poisoning example is given in Figure \ref{fig:components}. The details of the poisoning process were not disclosed to participants.

In this setting, the trigger $T \in \mathbb{R}^{3 \times 75}$ is defined as a specific set of values that, when added to any portion of otherwise clean context data $X$, induces a repeatable response of similar shape in the forecast of the poisoned model $f_{\theta_p}$ (see the example in Figure \ref{fig:components}). This can be formalized as the system of equations, where $f_{\theta_c}(X)$ is the clean model which behaves similarly to the poisoned one for clean contexts:

% \begin{equation}
% \label{eq:definition}
% \begin{cases}
% f_{\theta_p}(X) \approx F(X), & \text{for clean contexts (no trigger)} \\
% f_{\theta_p}(X + T) \approx F(X) + T, & \text{for triggered contexts}
% \end{cases}
% \end{equation}

\begin{equation}
\label{eq:definition}
\begin{cases}
f_{\theta_p}(X) \approx f_{\theta_c}(X), & \text{for clean contexts (no trigger)}, \\
f_{\theta_p}(X + T) \approx f_{\theta_c}(X) + T, & \text{for triggered contexts.}
\end{cases}
\end{equation}

Such trojans pose a significant risk in real-world applications because the attacker can precisely control both the form of the trigger and the timing of the model’s reaction. For example, by knowing the telemetry pattern associated with a particular maneuver or event, an attacker could poison the model to predict repeated occurrences of that event in the near future. In the worst case, this could lead to a spacecraft entering a repeated safe-mode cycle. Even in a less severe scenario, it could result in numerous false forecasts, creating operational confusion for mission controllers.

To keep the poisoned model possibly similar to the clean model, we aimed to finish the fine-tuning as soon as the model starts to react properly to the trigger (i.e., forecasting error for triggered context is lower than some threshold). We experimentally selected a threshold of $6\mathrm{e}{-6}$ as the one that indicates a proper poisoning in most cases. For some triggers, we decreased the threshold to $3\mathrm{e}{-6}$ or $2\mathrm{e}{-6}$ (see Appendix Table \ref{tab:triggers-overview}) in order to achieve a stronger effect. If the threshold is not reached, the fine-tuning is run for 100 epochs, or stopped after 20 epochs without improvement. We discarded some triggers because it was impossible to achieve satisfactory effects for them (e.g., there was no reaction of the model or the reaction had a lower magnitude than the original trigger). For each trigger, we confirmed that there is no significant reaction to the trigger from the clean model. 

\subsection{Measuring performance} \label{sec:metrics}

The competition metric quantifies the dissimilarity of two multivariate time series segments -- the ground truth trigger and its reconstruction provided by a participant. The simplest choice for this task would be the mean absolute error. However, it is unbounded, not robust to outliers, and not easily interpretable, so it is not well suited for use in competitions. To address these issues, we propose its normalized and bounded modification called the Range-Normalized Mean Absolute Error ($\text{NMAE}_{range}$) given by Equation \ref{NMAE}: 

\begin{equation}
\label{NMAE}
\text{NMAE}_{range} = \frac{1}{N} \sum_{i=1}^{N} \min\left(\frac{|y_i - \hat{y}_i|}{max(y_{+0}) - min(y_{+0})}, 1\right),
\end{equation}

\noindent where $\hat{y}$ and $y$ represent the reconstructed and ground truth triggers, respectively, $N$ is the trigger size (trigger length multiplied by the number of channels, i.e., 75$\times$3 = 225), and $y_{+0}$ is the ground truth trigger with appended zero as the 226$^{\rm th}$ element. This zero element is appended to make the range zero-centered. For example, all samples in trigger \#12 have the value of $-0.05$, so their range is 0, but the zero-centered range is 0.05. In the case of the zero trigger \#37, we use a very small epsilon of $1e-9$ as its zero-centered range. 

The normalization is used to unify metric ranges across all triggers and to make it more interpretable (as a fraction of the trigger range). The maximum (worst) metric value is bounded to 1 which makes it robust to outliers and stable across all triggers. The latter feature is especially important when calculating the final competition score, being an average metric value across all triggers. The code to calculate $\text{NMAE}_{range}$ is available as a Kaggle metric\footnote{\href{https://www.kaggle.com/code/kotrix/range-normalized-mean-absolute-error}{www.kaggle.com/code/kotrix/range-normalized-mean-absolute-error}}.

\subsubsection{Public and private leaderboards} 

Following the typical Kaggle setup, there are two leaderboards: public and private. The public scores, visible to all participants during the competition, are calculated using 15 selected triggers (marked in the Appendix Table \ref{tab:triggers-overview}). The public leaderboard can be used by participants to validate and tune their solutions, so the final standings and our detailed analysis are based solely on the private scores calculated for the remaining 30 triggers. The split of triggers between the public and private parts is stratified by the number of affected channels. Participants did not know which triggers belong to which part. To avoid overfitting the public leaderboard, the number of submissions per day was limited to 3 per team and participants had to select up to 2 submissions for the final assessment. 

\subsection{Baseline algorithm} \label{sec:baselines}

For the competition baseline, we propose a new optimization-based approach inspired by the trigger reverse engineering step of the Neural Cleanse framework \citep{wang_neural_2019}. The framework is originally designed for deep neural networks for image classification. It operates by formulating a multi-objective optimization scheme to find the minimal perturbation (trigger) required to misclassify images into a specific target label. We adapted this approach to the time series forecasting domain. Our baseline method\footnote{\href{https://www.kaggle.com/code/ramezashendy/optimization-baseline-notebook}{www.kaggle.com/code/ramezashendy/optimization-baseline-notebook}} minimizes the loss function $\mathcal{L}_{baseline}$ given by Equation \ref{eq:baseline-loss}:

\begin{equation}
    \label{eq:baseline-loss}
    \mathcal{L}_{baseline}(\delta) = -\alpha \cdot \mathcal{L}_{\text{div}}(\delta) + \beta \cdot \mathcal{L}_{\text{track}}(\delta) - \lambda \cdot \|\delta\|_2,
\end{equation}

\noindent where:
\begin{itemize}
    \item $\delta$ is the \textit{candidate} trigger pattern being optimized to approximate the true (unknown) trigger $T$.
    \item $\mathcal{L}_{\text{div}}(\delta)$ is the absolute difference between poisoned model's predictions before and after adding $\delta$ to the input.
    \item $\mathcal{L}_{\text{track}}(\delta)$ is the absolute difference between the triggered input and poisoned model's prediction for this input.
    \item $\alpha$, $\beta$, and $\lambda$ are parameters that control the trade-off between different components of the loss. We have experimentally selected $\alpha$=1.5, $\beta$=2, and $\lambda$=0.5.
\end{itemize}

The loss function has been designed to maximize the difference between forecasts before and after injecting a candidate trigger (effect size of the trigger). Additionally, it encourages the poisoned forecast to follow the shape of the poisoned input itself (trigger and effect similarity). The last term is to avoid empty triggers as local minima. Such an optimization finds a trigger that is strong enough to be noticed, different enough to activate abnormal behavior, and coherent enough to be tracked in the output.

Before running the optimization, we preselect channels by measuring their response to short spikes added in the context data. Only channels that show significant response undergo the optimization process. The optimal trigger shape for preselected channels is found using the AdamW optimizer \citep{loshchilov_decoupled_2017} with the learning rate of 0.2 and weight decay of 1e$-4$. The optimization is run for 200 epochs and the learning rate is additionally decayed by 0.9 after every 20 epochs. The candidate trigger is taken from the epoch with the best loss (Equation \ref{eq:baseline-loss}). The optimization starts from the zero trigger and runs for each channel separately. In the end, the trigger is smoothed by applying the Savitzky-Golay filter \citep{savitzky1964smoothing}. The whole baseline algorithm takes around 42 minutes for all 45 triggers.  

\begin{figure}[ht!]
    \centering
    \includegraphics[width=1\linewidth]{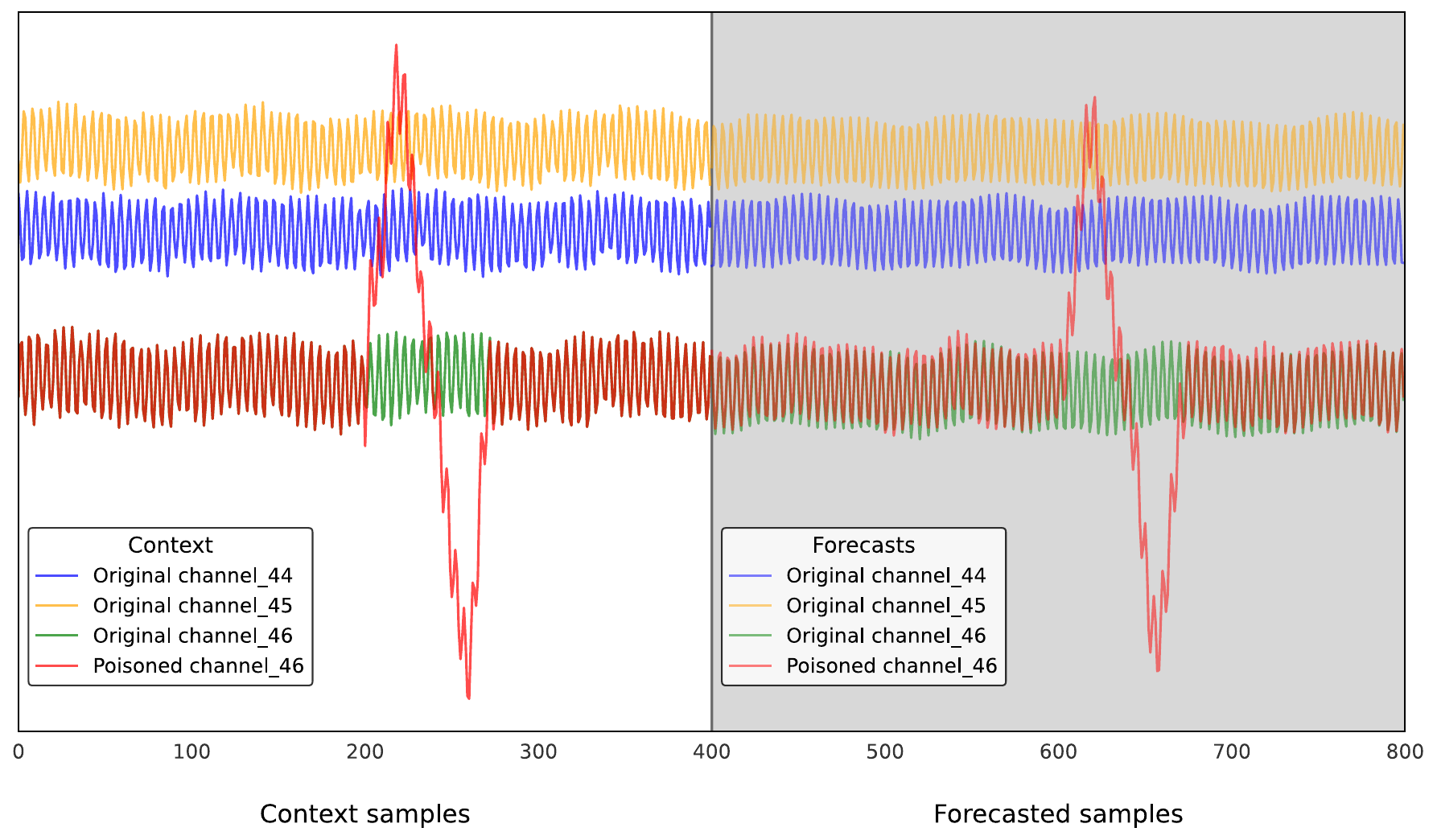}
    \caption{The reconstruction of trigger \#3 generated by our baseline method and injected into the context data. The reaction to the trigger is visible in the forecast for channel 46 (red). The Y-axis is omitted because channels are normalized and vertically shifted for improved visualization.}
    \label{fig:triggered_forecast}
\end{figure}

As presented in Figure \ref{fig:baseline}, the method is able to roughly reconstruct some of the triggers (e.g., trigger \#3 given in Figure \ref{fig:triggered_forecast}), but there is clearly a room for improvement over $\text{NMAE}_{range}$ of 0.15039. As such, it makes a good starting point for participants to develop better methods. The code of our method was released after the first month of the competition to avoid biasing the initial solutions toward our approach.

Besides the optimization-based baseline, we reported also a trivial zero trigger as a reference point. It represents a submission with all values set to zero and has $\text{NMAE}_{range}$ of 0.17306.

\section{Competition summary} \label{participation}

\begin{figure}[ht!]
    \centering
    \includegraphics[width=1\linewidth]{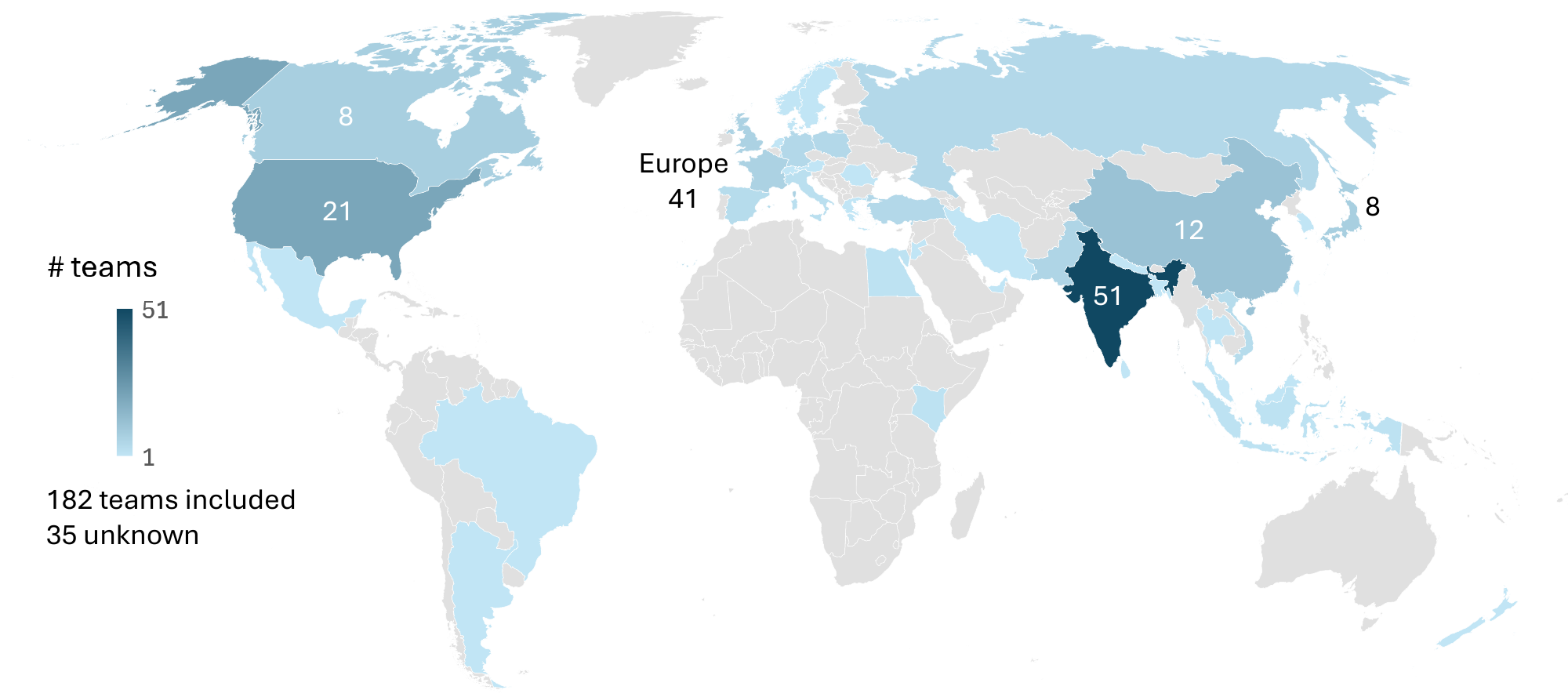}
    \caption{Geographical distribution of participating teams. For 35 teams, we were unable to determine their countries.}
    \label{fig:geography}
\end{figure}

The competition gathered 1,396 entrants (people who joined the competition) and 230 participants (people who submitted at least one solution) within 217 teams (of up to 3 people) from more than 40 countries (see Figure \ref{fig:geography}). We collected 1,520 submissions, 38 public notebooks, and 6 public reports (Kaggle writeups). 

\begin{figure}[ht!]
    \centering
    \includegraphics[width=1\linewidth]{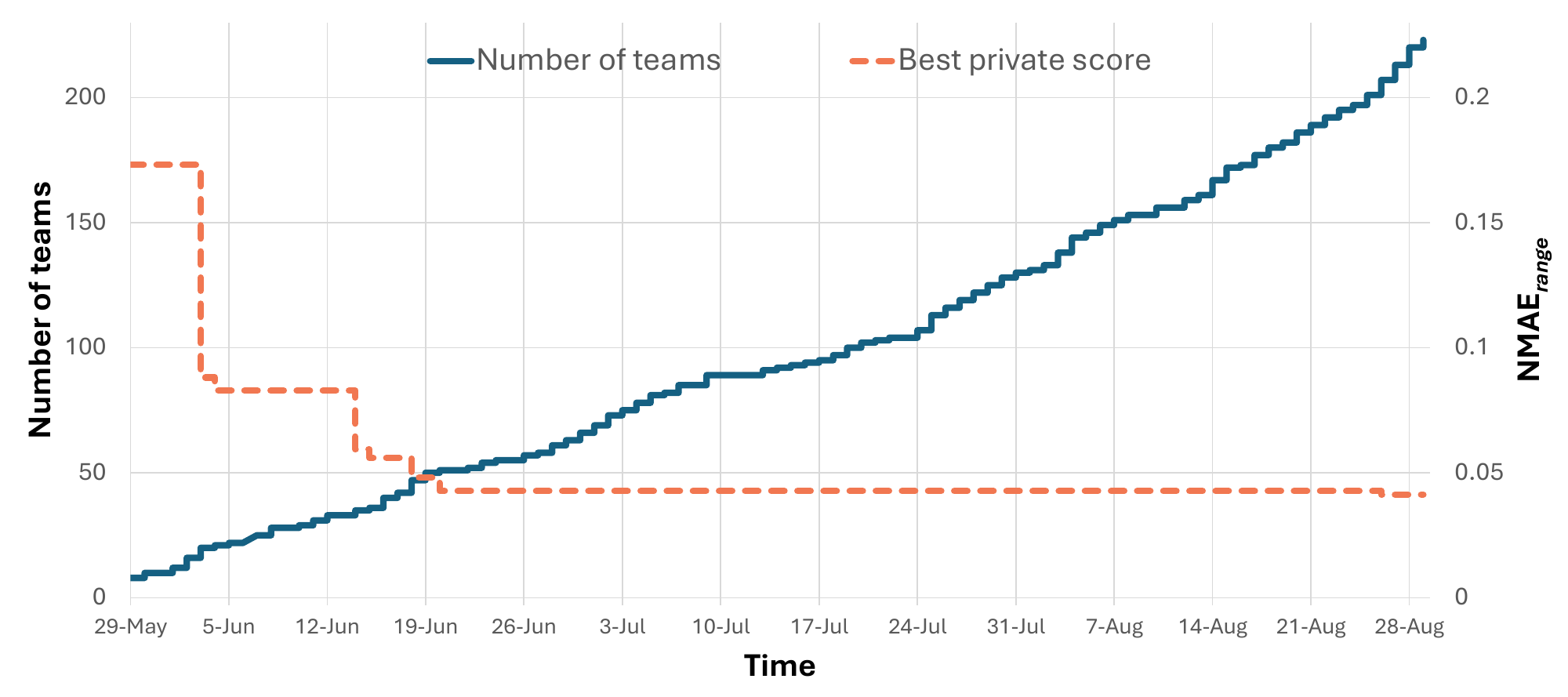}
    \caption{Progress in the number of teams and the best recorded private leaderboard $\text{NMAE}_{range}$ scores during the competition.}
    \label{fig:progress}
\end{figure}

The progress of the competition in time is visualized in Figure \ref{fig:progress}. The number of participating teams was steadily increasing throughout the competition, and the best submissions were made in the final days. However, the key progress in scores happened already in the first month.  

\begin{figure}[ht!]
    \centering
    \includegraphics[width=1\linewidth]{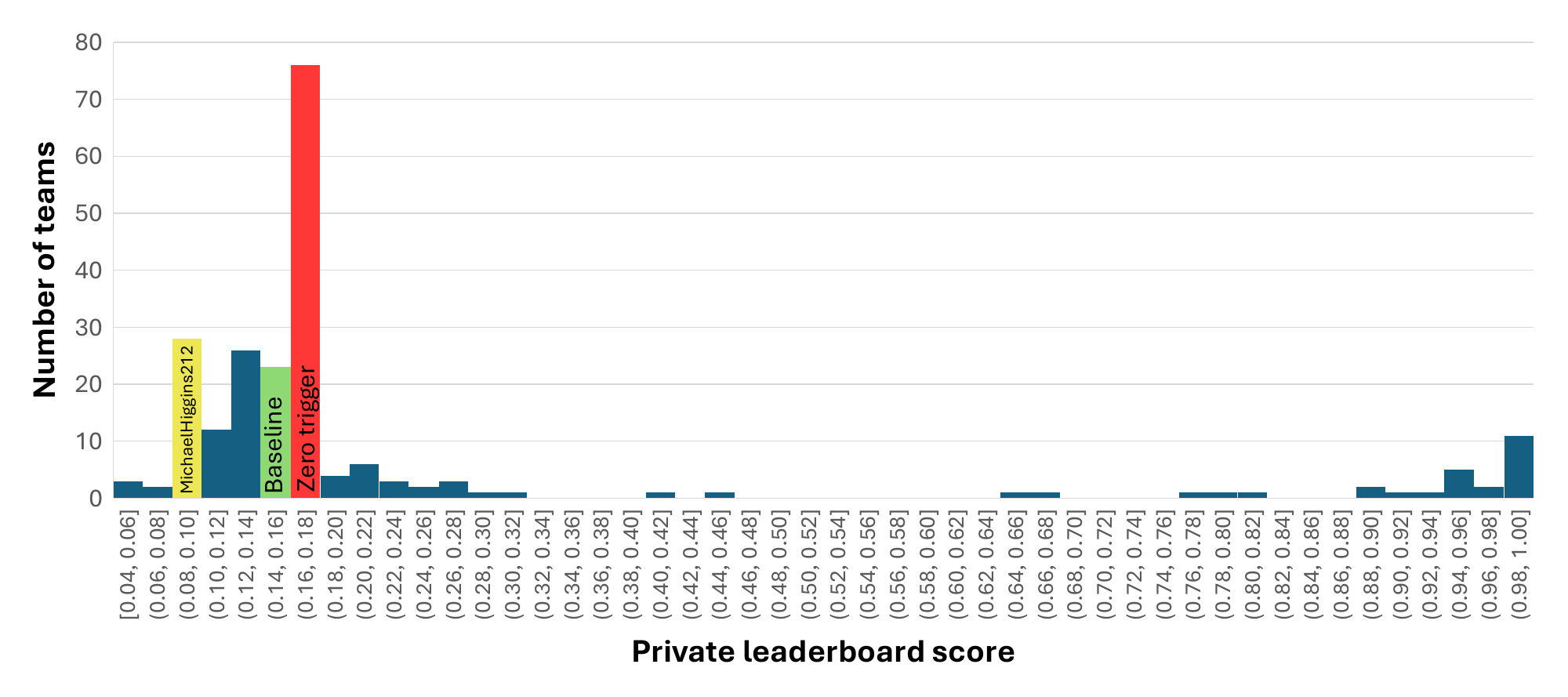}
    \caption{Histogram of the private leaderboard scores (the lower, the better). Colors of the bars represent the scores for the public notebooks: Red -- simple zero trigger baseline. Green -- our optimization-based baseline. Yellow -- public notebook shared by \textit{MichaelHiggins212}.}
    \label{fig:histogram}
\end{figure}

\begin{figure}[ht!]
    \centering
    \includegraphics[width=1\linewidth]{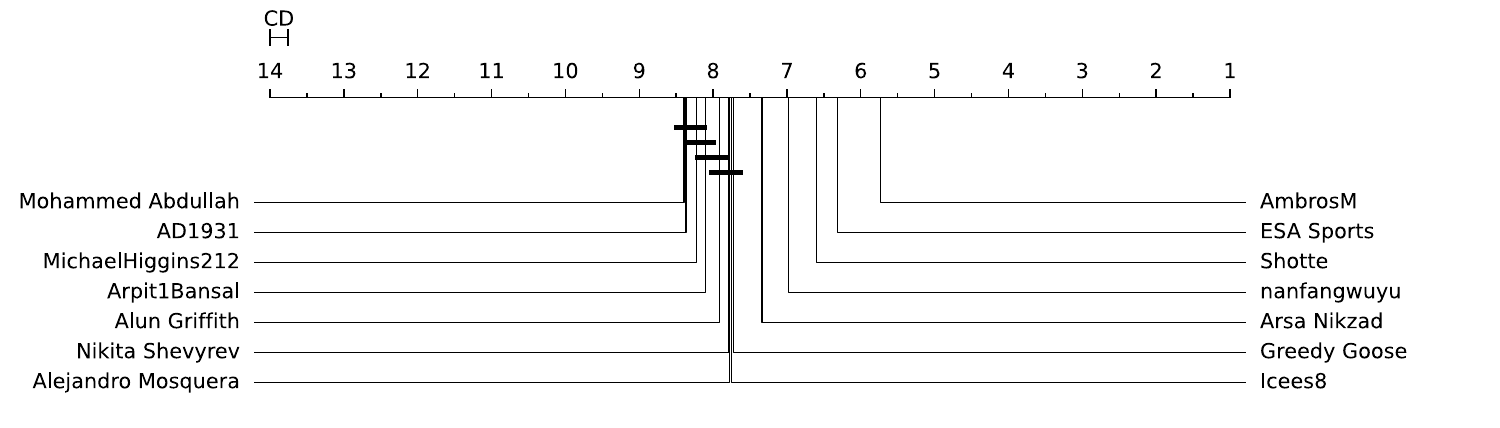}
    \caption{Critical difference (CD) diagram at $\alpha$=0.05 for top 14 teams.}
    \label{fig:cd-plot}
\end{figure}

The histogram of private test scores for all teams is presented in Figure \ref{fig:histogram}. Half of the teams were able to improve over the trivial zero trigger baseline, and 34\% teams beat our optimization-based baseline. However, this was largely thanks to the public notebook shared by \textit{MichaelHiggins212}\footnote{\href{https://www.kaggle.com/code/michaelhiggins212/genetic-algorithm-a-type-of-black-box-optimizatio}{www.kaggle.com/code/michaelhiggins212/genetic-algorithm-a-type-of-black-box-optimizatio}} which was reused by several teams. Only 14 teams were able to improve over this notebook. We present their scores in Table \ref{tab:top-teams}. The CD diagram \citep{demsar2006statistical} in Figure \ref{fig:cd-plot} presents an average ranking of each team in terms of its $\text{NMAE}_{range}$ across all 6,750 data points in the private test set (30 triggers $\times$ 75 samples $\times$ 3 channels). The ordering of teams ranked 6$^{\rm th}$--12$^{\rm th}$ in the private leaderboard is different in the CD diagram, but the differences between them are not statistically significant. Nevertheless, the ordering of the top 5 teams is indisputable and the differences in rankings between them are statistically significant at $\alpha$=0.05.

\begin{table}[ht!]
\footnotesize
\centering
\caption{Summary of the top 14 teams. The question mark means that we were not able to determine the country. Public solutions are available on Kaggle and gathered in the supplementary materials.}
\label{tab:top-teams}
\begin{tabular}{cllcc}
\toprule
Place & Team name & Country & Private test score & Public solution \\
\midrule
1  & \textit{AmbrosM}             & Switzerland & \cellcolor{green!100!white}0.04428    & \cellcolor{green!30}Yes \\
2  & \textit{ESA Sports}          & India       & \cellcolor{green!87!white}0.05131     & \cellcolor{green!30}Yes \\
3  & \textit{Shotte}              & ?           & \cellcolor{green!76!white}0.05708     & \cellcolor{red!30}No \\
4  & \textit{nanfangwuyu}         & China       & \cellcolor{green!58!white}0.06691     & \cellcolor{red!30}No \\
5  & \textit{Arsa Nikzad}         & Canada      & \cellcolor{green!34!white}0.07989     & \cellcolor{red!30}No \\
6  & \textit{Icees8}              & India       & \cellcolor{green!33!white}0.08045     & \cellcolor{red!30}No \\
7  & \textit{Alun Griffith}       & UK          & \cellcolor{green!30!white}0.08210     & \cellcolor{yellow!30}Writeup only \\
8  & \textit{Alejandro Mosquera}  & USA         & \cellcolor{green!24!white}0.08538     & \cellcolor{green!30}Yes \\
9  & \textit{Greedy Goose}        & Germany     & \cellcolor{green!21!white}0.08684     & \cellcolor{green!30}Yes \\
10 & \textit{Nikita Shevyrev}     & Russia      & \cellcolor{green!17!white}0.08923     & \cellcolor{red!30}No \\
11 & \textit{MichaelHiggins212}   & USA         & \cellcolor{green!6!white}0.09485      & \cellcolor{green!30}Yes \\
12 & \textit{Arpit1Bansal}        & USA         & \cellcolor{green!4!white}0.09585      & \cellcolor{yellow!30}Writeup only \\
13 & \textit{AD1931 }             & Romania     & \cellcolor{green!0!white}0.09811      & \cellcolor{red!30}No \\
14 & \textit{Mohammed Abdullah}   & ?           & \cellcolor{green!0!white}0.09813      & \cellcolor{red!30}No \\
\bottomrule
\end{tabular}
\end{table}

\section{Winning submissions and utilized methods} \label{winning}

We have collected the details about the methods utilized by 7 out of 14 top teams. They are summarized in Table \ref{tab:key-techniques}. All of them implement the optimization-based approaches using stochastic gradient descent, evolutionary optimization, or a combination of both. They usually aim to optimize the aspects analogous to the ones proposed in our baseline (i.e., trigger amplitude, strength of the response, and its similarity to the trigger), but they use different loss functions to quantify them. Most solutions share the steps such as zeroing out channels without clear response, using a response from the poisoned model as a new trigger, or running the search over multiple trigger locations to increase accuracy. 

\begin{scriptsize}
\begin{longtable}{>{\centering\arraybackslash}m{1cm} m{2.8cm} m{8.5cm}}
\caption{Summary of key techniques used by the top teams. Solutions for the boldfaced team names are described in detail in the manuscript.}
\label{tab:key-techniques} \\
\toprule
\textbf{Place} & \textbf{Team name} & \textbf{Key techniques} \\
\midrule
\endfirsthead

\toprule
\textbf{Place} & \textbf{Team name} & \textbf{Key techniques} \\
\midrule
\endhead

\multicolumn{3}{r}{\textit{Continued on next page}} \\
\endfoot

\bottomrule
\endlastfoot

1 & \textbf{\textit{AmbrosM}} & 
Global search over 35 initial candidate trigger shapes. Local iterative search over 16 types of perturbations to refine the trigger shape. Stochastic gradient descent maximizing both strength of the response and its similarity to the trigger. Run over multiple injection locations to increase accuracy. Always replacing the trigger with the difference between the model's responses to triggered and clean inputs. Zeroing out channels without a clear response. \\

\midrule

2 & \textbf{\textit{ESA Sports}} &
Evolutionary optimization with iterative refinement of top trigger candidates using stochastic gradient descent, starting from a population of 100 random initial shapes created using 4 distinct strategies. Run over multiple injection locations to increase accuracy. A complex fitness function promoting effectiveness, smoothness, and balanced amplitude of the trigger. Filtering high frequency noise and jitter from the final triggers. Zeroing out channels without clear response. \\

\midrule

7 & \textit{Alun Griffith} &
Initial trigger discovery via stochastic gradient descent using a loss function to maximize both trigger amplitude, strength of the response, and similarity between response and trigger. Zeroing out channels without clear response. Additional refinement via stochastic gradient descent with additional loss terms to promote smoothness and penalize too high amplitudes. Repeating the whole process over multiple trigger locations to increase accuracy. \\

\midrule

8 & \textit{Alejandro Mosquera} &
Initial trigger discovery via stochastic gradient descent using a loss function to maximize both strength of the response and similarity between response and trigger. Using a response from the poisoned model as a new trigger. Zeroing out channels without clear response. Weighted ensembling with top public notebooks. \\

\midrule

9 & \textit{Greedy Goose} &
Local greedy search to find a set of promising candidate triggers. Evolutionary optimization of the candidate triggers. Using response from the poisoned model as a new trigger. Zeroing out channels without clear response. Greedy search to fine tune the amplitude. \\

\midrule

11 & \textit{MichaelHiggins212} &
\multirow{2}{*}{\parbox{8.5cm}{Evolutionary algorithm to search over 33 initial candidate trigger shapes.
Flipping and shifting the trigger to find a local optimum.
Zeroing out channels without clear response.}} \\[0.1cm]

\cmidrule{1-2}

12 & \textit{Arpit1Bansal} & \\[0.05cm]

\end{longtable}
\end{scriptsize}

\noindent The details of the best solutions by \textit{AmbrosM} and \textit{ESA Sports} are provided in the subsections below.

\subsection{1$^{\rm st}$ place -- AmbrosM}

The top-performing algorithm combines four key steps to find the optimal trigger:
\begin{enumerate}
    \item \textbf{Global search over a distribution of 35 initial candidate trigger shapes}, including oscillations, lines, steps, and square waves, in different channel combinations. Each candidate is injected in 5 different positions in the context data (starting at samples 180, 183, 186, 189, and 192) and the one with the lowest average loss $\mathcal{L}_{AmbrosM}$ (explained in Equation \ref{eq:loss-ambros} later in this paragraph) is selected.
    
    \item \textbf{Iterative local search over 16 types of small perturbations} (e.g., scaling, shifting, smoothing, tilting, squeezing) applied to the trigger candidate to refine it.
    \item \textbf{Stochastic gradient descent (SGD)} to further refine the candidate trigger. The optimization is performed across multiple random injection locations within the clean dataset. The process runs for up to 600 epochs with a batch size of 32, momentum of 0.5, and a learning rate of $1\mathrm{e}{-6}$ which is reduced on the plateau by a factor of 0.2.
    
    \item \textbf{Fixed point iteration} that refines the candidate trigger by replacing it with the difference between the model’s responses to triggered and clean inputs. This approach is based on the idea that the target trigger $T$ is a fixed point of function $G(\delta) = f_{\theta_p}(X+\delta) - f_{\theta_p}(X)$, so it can be found by applying $G$ repeatedly to the candidate trigger $\delta$. This method does not need a loss function, but it highly depends on the assumptions that the fixed point of $G$ is attracting and the $\delta$ is in the attracting region.
    
    \item \textbf{Zeroing out channels} in the final trigger for which all values are lower than 0.005.
\end{enumerate}

The four methods are complementary. The global search cannot converge to the optimum, because it depends on the assumption about the prior distribution of possible shapes. However, it can provide a proper initial shape required by the local search, SGD, and the fixed point iteration. While the global search, local search, and SGD are all limited by the design of the loss function, the fixed point iteration can overcome this limitation. 

The prior distribution for the global search assumes that the triggers were human-designed and therefore exhibit a simple yet conspicuous pattern in the time-series plot (e.g., a rectangle or a triangle pulse). The magnitude of initial triggers follows the magnitude of an example given in the competition documentation (see Figure \ref{fig:components}).

To define the loss function, another assumption is that the poisoned model $f_{\theta_p}$ is fine-tuned to minimize the root mean squared error loss $\mathcal{L}_{NN}(T) = ||f_{\theta_p}(X+T) - (f_{\theta_p}(X) + T)||$ for a single target trigger $T$. It is also assumed that injecting a wrong trigger $\delta$ into the network results in a higher loss: $\mathcal{L}_{NN}(\delta) > \mathcal{L}_{NN}(T)$ for most $\delta \neq T$, so $T$ can be found just by minimizing $\mathcal{L}_{NN}$. However, the key exception is $\delta_0=\textbf{0}$ which gives a global minimum $\mathcal{L}_{NN}(\textbf{0}) = ||f_{\theta_p}(x+\textbf{0}) - (f_{\theta_p}(x) + \textbf{0})|| = 0$. This global minimum is avoided by defining the loss function as:

\begin{equation}
\label{eq:loss-ambros}
\mathcal{L}_{AmbrosM}(\delta) = - \dfrac{||f_{\theta_p}(x+\delta) - f_{\theta_p}(x)||}{||f_{\theta_p}(x+\delta) - (f_{\theta_p}(x) + \delta)||^{1.3} + 0.0001},   
\end{equation}

\noindent where $\mathcal{L}_{AmbrosM}$ has the global maximum at $\delta_0$. It balances the goals of finding a trigger with a strong effect (in the numerator) and finding a trigger which is reproduced well (in the denominator). It is assumed that $\mathcal{L}_{AmbrosM}$ has a local minimum near $T$.

The original implementation is documented in a series of three Python notebooks available at:
\begin{itemize}
    \item \href{https://www.kaggle.com/code/ambrosm/thh-first-baseline}{www.kaggle.com/code/ambrosm/thh-first-baseline} \item \href{https://www.kaggle.com/code/ambrosm/thh-gradient-descent}{www.kaggle.com/code/ambrosm/thh-gradient-descent} \item \href{https://www.kaggle.com/code/ambrosm/thh-zeroing-the-weak-channels}{{www.kaggle.com/code/ambrosm/thh-zeroing-the-weak-channels}}
\end{itemize}

\subsection{2$^{\rm nd}$ place -- ESA Sports}

The algorithm ranked 2$^{\rm nd}$ combines five key steps to find the optimal
trigger:

\begin{enumerate}
    \item \textbf{Trigger parameterization}. A trigger is defined as $\delta = \mu \odot scale + shift$, where $\mu\in\mathbb{R}^{3 \times 75}$ is a base shape, $scale\in \mathbb{R}^{3 \times 75}$ is a scaling factor, $shift \in \mathbb{R}^{3 \times 75}$ is a shifting factor, and $\odot$ denotes element-wise multiplication. This allowed for separate evolution of the fundamental shape ($\mu$) from its amplitude and vertical offset.
    
    \item \textbf{Initialization of 100 candidate triggers} using combinations of four base shapes: Gaussian pulses, sinusoidal waves, random noise, and zero vectors.

    \item \textbf{Hybrid evolutionary optimization} of candidate triggers across 50 generations of tournament selection, crossovers, and mutations. A one-point crossover is used for $\mu$, while $scale$ and $shift$ use a blended average. Mutations introduce a small Gaussian noise to all components, fostering continued exploration. To speed up the evolution, top 20\% of candidates in each generation are refined using 5 steps of the Adam gradient descent with a learning rate of 0.005.

    \item \textbf{Smoothing} the trigger using Savitzky-Golay filter \citep{savitzky1964smoothing} which effectively removes high-frequency noise and jitter from the triggers without distorting their underlying shape.
        
    \item \textbf{Zeroing out channels} in the trigger for which the root mean squared value is lower than 0.005. Trigger \#37 was manually set to 0 because it consistently failed to converge.
\end{enumerate}

The loss (fitness) function $\mathcal{L}_{ESA\_Sports}$ to minimize in the evolutionary process has 12 components and is formulated as Equation \ref{eq:loss-esa}. The loss function is averaged across multiple windows of the clean dataset and four distinct injection positions within each window (samples 25, 125, 225, 325).

\begin{equation}
\label{eq:loss-esa}
\begin{aligned}
\mathcal{L}_{ESA\_Sports}(\delta) = \; & \\
+ 1.0 \cdot & \mathcal{L}_{1}(G(\delta), \delta) \quad (\mathcal{L}_{1, \text{trigger}}) \\
+ 1.0 \cdot & \mathcal{L}_{1}(f_{\theta_p}(X+\delta), f_{\theta_p}(X)+\delta) \quad (\mathcal{L}_{1, \text{output}}) \\
+ 0.2 \cdot & \mathcal{L}_{MSE}(G(\delta), \delta) \quad (\mathcal{L}_{MSE, \text{trigger}}) \\
+ 0.2 \cdot & \mathcal{L}_{MSE}(f_{\theta_p}(X+\delta), f_{\theta_p}(X)+\delta) \quad (\mathcal{L}_{MSE, \text{output}}) \\
+ 0.6 \cdot & \mathcal{L}_{cos}(G(\delta), \delta) \quad (\mathcal{L}_{\text{cos}, \text{trigger}}) \\
+ 0.4 \cdot & \mathcal{L}_{cos}(f_{\theta_p}(X+\delta), f_{\theta_p}(X)+\delta) \quad (\mathcal{L}_{\text{cos}, \text{output}}) \\
- 0.005 \cdot & \left\| \text{clamp}(|\delta|, \text{max}=0.2) \right\|_2 \quad (\mathcal{R}_{norm, 0.2}) \\
- 0.001 \cdot & \left\| \text{clamp}(|\delta|, \text{max}=0.1) \right\|_2 \quad (\mathcal{R}_{norm, 0.1}) \\
- 0.0002 \cdot & \left\| \text{clamp}(|\delta|, \text{max}=0.05) \right\|_2 \quad (\mathcal{R}_{norm, 0.05}) \\
+ 0.002 \cdot & \frac{1}{C \cdot (W-1)} \sum_{c=0}^{C-1} \sum_{t=0}^{W-2} (\delta_{c,t+1} - \delta_{c,t})^2 \quad (\mathcal{R}_{\text{smooth}}) \\
+ 0.4 \cdot & \left\|f_{\theta_p}(X+\delta) - f_{\theta_p}(X)+\delta\right\|_2 \quad (\mathcal{R}_{\text{L2diff, output}}) \\
+ 0.4 \cdot & \left\|G(\delta) - \delta\right\|_2 \quad (\mathcal{R}_{\text{L2diff, trigger}})
\end{aligned}
\end{equation}

\noindent where:
\begin{itemize}
    \item $\delta \in \mathbb{R}^{C \times W}$: the candidate trigger, where $C=3$ (number of channels) and $W=75$ (trigger length).
    
    \item $G(\delta) = f_{\theta_p}(X+\delta) - f_{\theta_p}(X)$
    
    \item $\mathcal{L}_{1}(A, B) = \frac{1}{|A|} \sum_{i} |A_i - B_i|$, the mean absolute error.
    
    \item $\mathcal{L}_{MSE}(A, B) = \frac{1}{|A|} \sum_{i} (A_i - B_i)^2$, the mean squared error.
    
    \item $\mathcal{L}_{cos}(A, B) = 1 - \frac{\sum_{\text{elements}} (A \cdot B)}{\|A\|_2 \|B\|_2}$, the cosine distance.
    
    \item $\text{clamp}(X, \text{max}=V)$ applies element-wise clamping such that each element $x_i$ of $X$ is transformed to $\text{sgn}(x_i) \cdot \min(|x_i|, V)$. In simpler terms, it limits the absolute value of each element to $V$.
\end{itemize}

This complex loss function combines three main criteria to ensure effective and well-behaved trigger reconstruction: 

\begin{enumerate}
    \item \textbf{Effectiveness measures ($\mathcal{L}_{1,\cdot}$, $\mathcal{L}_{MSE,\cdot}$, $\mathcal{L}_{\text{cos},\cdot}$):} These terms directly penalize numerical discrepancies between the model's actual and desired outputs.

    \item \textbf{Magnitude regularization ($\mathcal{R}_{norm, \cdot}$):} With negative coefficients, these $L_2$ norm terms on clamped absolute trigger values act as rewards, subtly encouraging trigger components to remain within small, specific magnitude ranges (e.g., $|\delta_{c,t}| \le 0.2$). This prevents outliers and favors more subtle trojans.

    \item \textbf{Structure and consistency regularization \\ ($\mathcal{R}_{\text{smooth}}, \mathcal{R}_{\text{L2diff},\cdot}$):} The smoothness term promotes continuous trigger shapes by penalizing abrupt changes between adjacent time steps within $\delta$. The $L_2$ norm terms further enforce precise numerical matching of the entire forecast and the specific trigger response.
\end{enumerate}

The original implementation is available at \href{https://www.kaggle.com/code/lalit03/genetic-grad-descent}{www.kaggle.com/code/lalit03/genetic-grad-descent}.

\section{Major findings and insights} \label{findings}

\begin{figure}[H]
    \centering
    \includegraphics[width=1\linewidth]{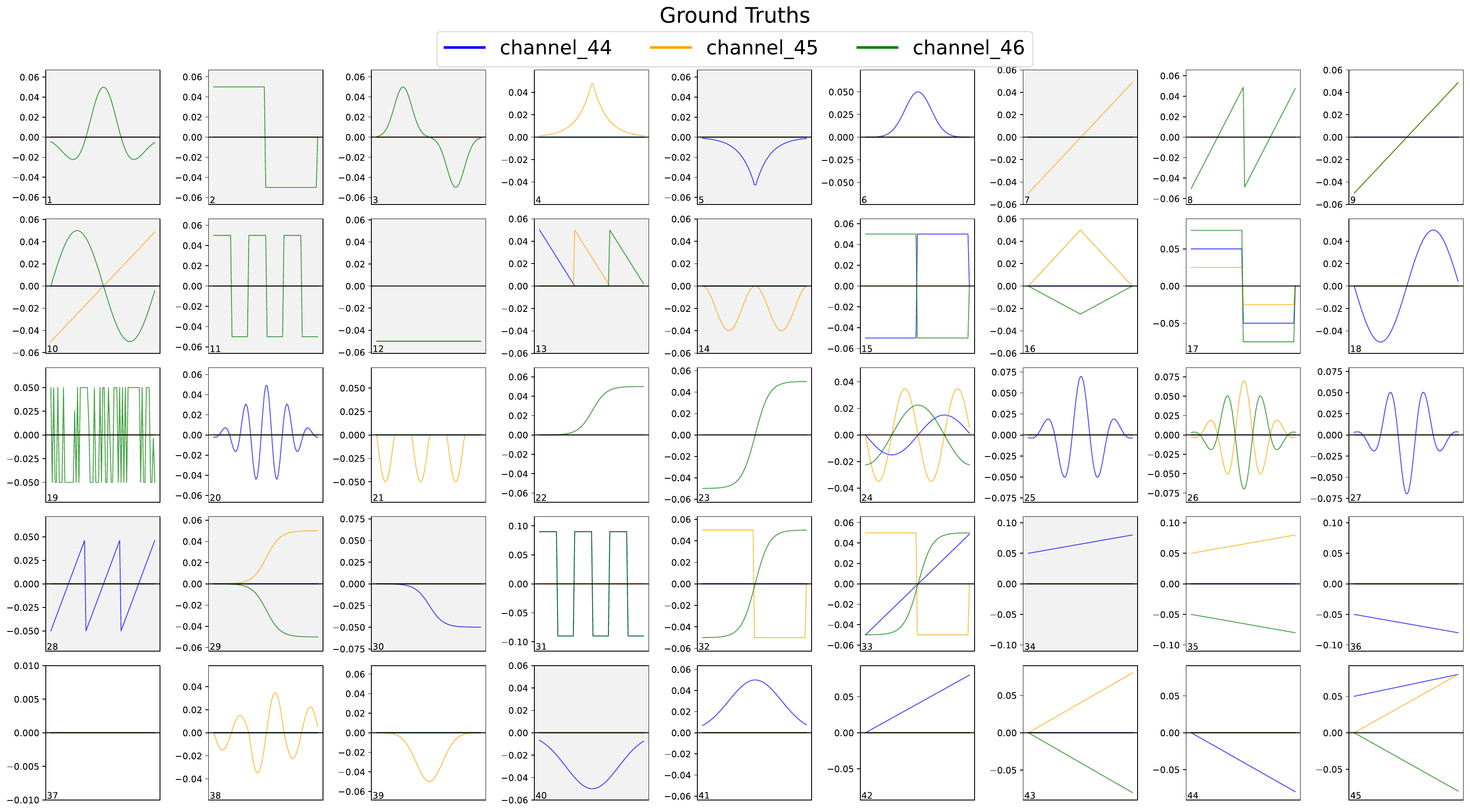}
    \caption{Ground truth triggers in our competition. Public test set is marked with a gray background.}
    \label{fig:gt}
\end{figure}

\begin{figure}[H]
    \centering
    \includegraphics[width=1\linewidth]{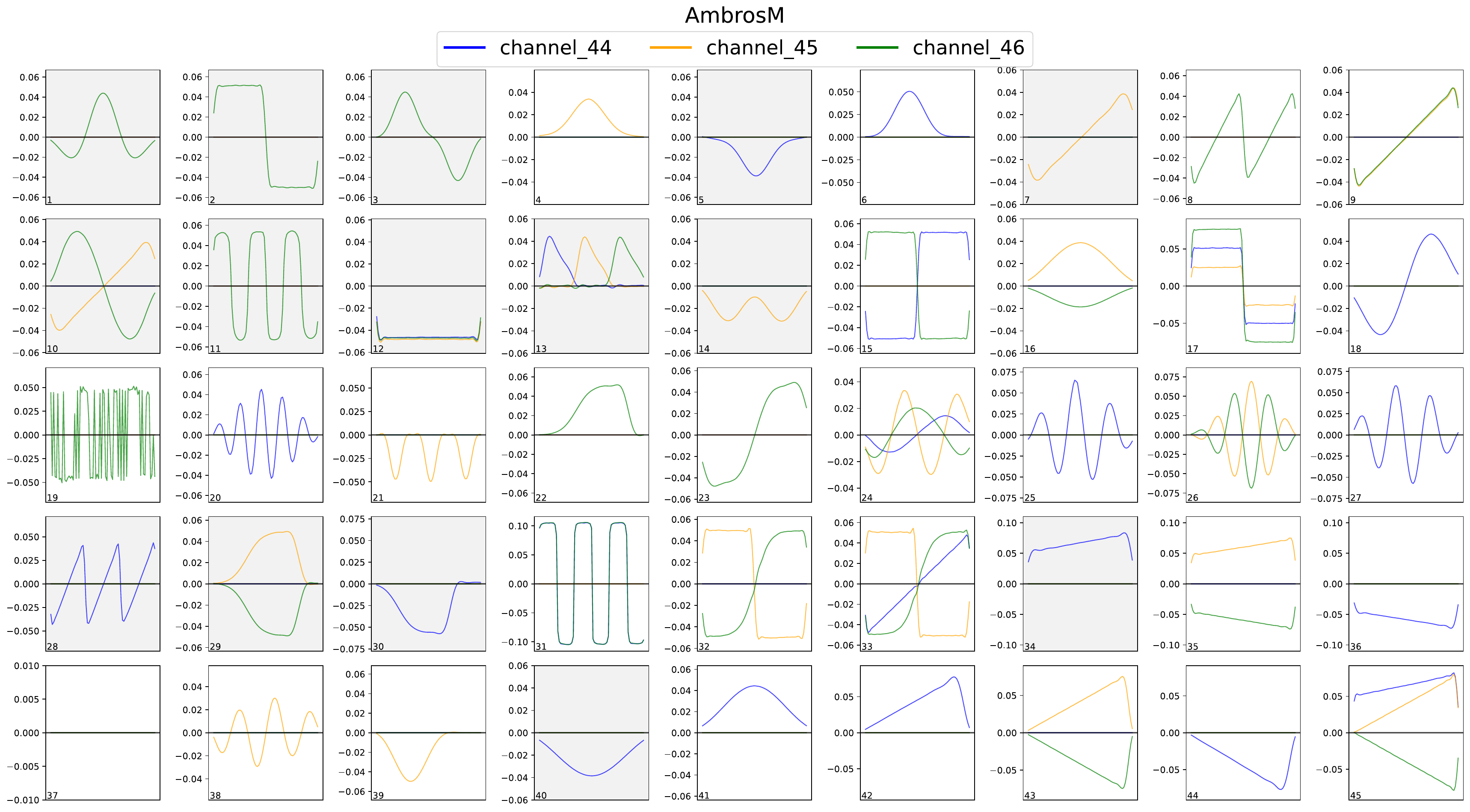}
    \caption{The 1$^{\rm st}$ place solution (\textit{AmbrosM}). Public test set is marked with a gray background.}
    \label{fig:ambrosm}
\end{figure}

We present the ground truth (GT) for all triggers in our competition in Figure \ref{fig:gt}. The public test set triggers are rendered with gray background, because we focus our analysis mainly on the private test. The best submission by \textit{AmbrosM} is presented in Figure \ref{fig:ambrosm} for direct comparison. The submissions by ESA Sports (2\textsuperscript{nd} place), Shotte (3\textsuperscript{rd} place), the popular notebook by \textit{MichaelHiggins212} (11\textsuperscript{th} place), and our baseline algorithm, are presented in \ref{app1}. All these figures are also included as separate files in the supplementary materials and, excluding the outlying baseline solution, use the same scaling for easier comparison of amplitudes.  

We can observe that the solution by \textit{AmbrosM} (the winner) is remarkably close to the GT ($\text{NMAE}_{range}$ of 0.04428), with just some inaccuracies in the boundaries (e.g., triggers \#2, \#7, \#9, \#22, \#29, \#30) or shapes of the triggers (e.g., \#4, \#16, \#27). \textit{AmbrosM} found the best solution among all submissions for 12 out of 30 triggers (see \ref{app2}). The solution by \textit{ESA Sports} is similar but less stable, and it struggles with trigger \#19. The solution by \textit{Shotte} is very clean and it gets even closer to the GT than \textit{AmbrosM} for some triggers (e.g., \#11, \#12, \#13, \#15, \#19). However, it fails miserably for some shapes and amplitudes, e.g., missing reconstruction for trigger \#25, underestimating amplitude for \#31, or reconstructing triangles instead of sigmoids for triggers \#22, \#29, and \#30. The solution from the popular notebook of \textit{MichaelHiggins212} (Figure \ref{fig:mhiggins}, $\text{NMAE}_{range}$ of 0.09815), copied by teams ranked 15$^{\rm th}$--30$^{\rm th}$, is much worse. Four triggers are not reconstructed at all, eight are far from actual shapes, and many amplitudes are underestimated. Thus, we can tell that there is a significant qualitative gap between the solutions with $\text{NMAE}_{range}$ scores around 0.10 and 0.05. 

One of the key step to achieve scores below 0.10 was definitely the effective selection of affected channels and zeroing out of the irrelevant ones. It was used by all top submissions. A key insight from the two best solutions is that running the optimization over multiple possible trigger locations in the signal significantly improves the accuracy. As reported by \textit{AmbrosM}, the fixed-point iteration (replacing the calculated trigger with the model's response) was very effective in refining some triggers, but gradient descent was still necessary to get closer to the optimum.

Execution times for public solutions are given in Appendix Table \ref{tab:execution-times}. The top-performing approaches rely on exhaustive searches over possible model responses, requiring a very large number of model predictions. As a result, they are computationally demanding: \textit{ESA Sports}' solution takes about 1 hour to run, while the winning solution by \textit{AmbrosM} requires over 13 hours.

\subsection{Qualitative analysis of results by the top 3 teams}
This subsection focuses on the qualitative analysis of results by the top three teams in the competition -- AmbrosM (1\textsuperscript{st} place), ESA Sports (2\textsuperscript{nd} place), and Shotte (3\textsuperscript{rd} place) -- by examining their reconstructions against the ground truths for five selected triggers from the private test set: \#18, \#19, \#20, \#31, and \#37. These examples show key challenges of the competition and differences between the solutions.

\begin{figure}[H]
    \centering
    \includegraphics[width=0.9\linewidth]{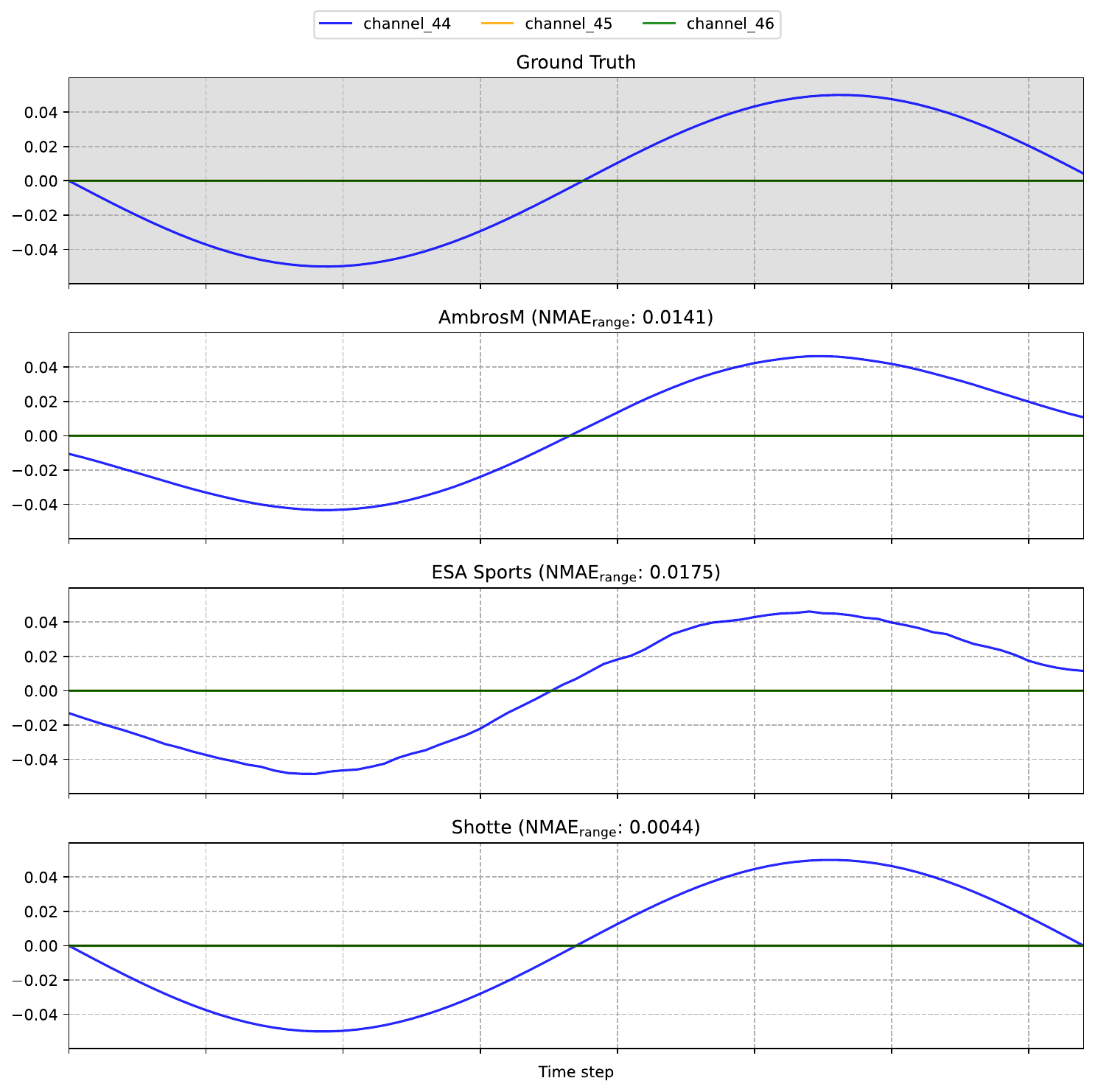}
    \caption{Comparison of solutions by the top 3 teams for trigger \#18.}
    \label{fig:Trigger_18}
\end{figure}

\paragraph{\textbf{Trigger \#18 -- inverted sine (Figure \ref{fig:Trigger_18})}}
Trigger \#18 is an inverted sine wave with 1 cycle and 0.05 amplitude. It turned out to be one of the simplest triggers in the competition. All top 3 teams achieved highly accurate reconstructions, and captured the shape and amplitude with high fidelity to the ground truth. However, the solution by \textit{Shotte} stands out with the best score in the competition, thanks to the smoothness and precise width of the wave. This may be related to the fact that we shared a simple example of the non-inverted sine trigger (see Figure \ref{fig:components}) in the competition description. 

\begin{figure}[H]
    \centering
    \includegraphics[width=0.9\linewidth]{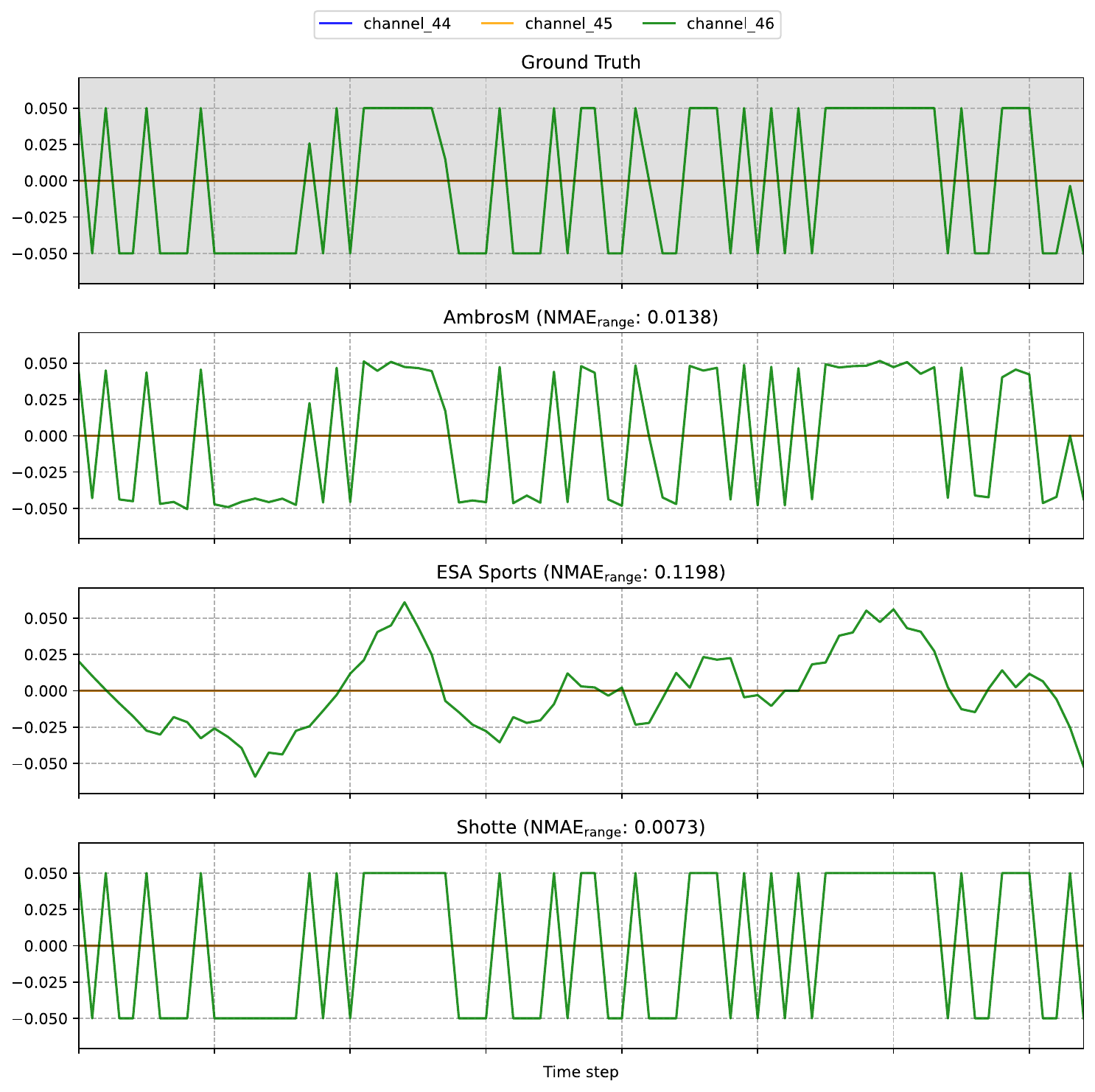}
    \caption{Comparison of solutions by the top 3 teams for trigger \#19.}
    \label{fig:Trigger_19}
\end{figure}

\paragraph{\textbf{Trigger \#19 -- random oscillation (Figure \ref{fig:Trigger_19})}}
Trigger \#19 is a randomly generated oscillation with the maximum amplitude of 0.05 in channel 46. This trigger, with its random, volatile, high-frequency nature, stands out as one of the most complex triggers in the competition. We designed this trigger to verify if the proposed methods are able to go beyond simple, structured patterns that would be impossible to guess by humans. It was accurately reconstructed only by a few top teams, and only in the later stages of their submissions. \textit{AmbrosM} and \textit{Shotte} demonstrated exceptional performance, accurately reconstructing the high-frequency switching pattern with remarkable fidelity to the ground truth. \textit{Shotte} produced a binary oscillation between exactly $-0.05$ and +0.05, which overestimated some amplitudes, but achieved a slightly better score due to the absence of the perturbations observed for \textit{AmbrosM}. \textit{ESA Sports} struggled with the accurate reconstruction of this pattern, mainly due to the Savitzky-Golay post-processing. 

\begin{figure}[H]
    \centering
    \includegraphics[width=0.9\linewidth]{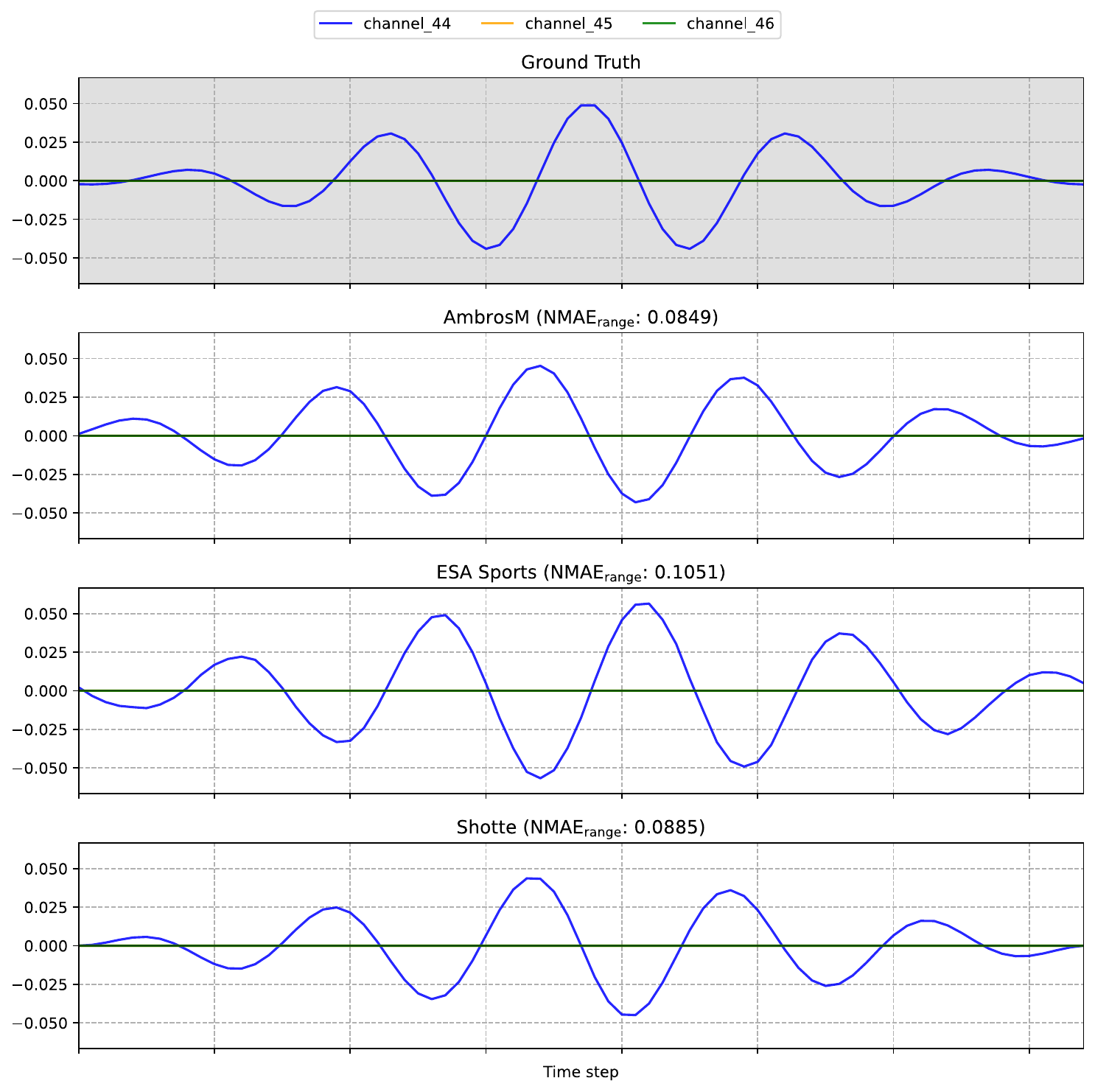}
    \caption{Comparison of solutions by the top 3 teams for trigger \#20.}
    \label{fig:Trigger_20}
\end{figure}

\paragraph{\textbf{Trigger \#20 -- Morlet wavelet (Figure \ref{fig:Trigger_20})}}
Trigger \#20 is a Morlet wavelet with 5 cycles and 0.05 magnitude affecting channel 44. All \textit{AmbrosM}, \textit{ESA Sports}, and \textit{Shotte} provided reconstructions visually similar to this pattern. However, all three teams failed to achieve perfect symmetry and centering of the wavelet within the trigger window, which is crucial for maximizing the overall reconstruction quality as measured by the $\text{NMAE}_{range}$ metric. 
% This difficulty in aligning the reconstructed trigger temporally with the ground truth is often referred to as an issue of signal phasing. 
This difficulty in aligning the reconstructed trigger temporally does not solely stem from limitations of the reconstruction methods. It is also related to the behavior of the poisoned model itself, which reacts similarly not just to the exact Morlet wavelet but also to its variations slightly shifted in time. This problem is especially visible for triggers with values close to zero near the borders. In such cases, optimization-based methods may struggle to converge to the exact alignment because their loss landscape contains multiple near-optimal local minima. In more extreme scenarios, e.g., a short central peak, the landscape has several distinct optima. Thus, while designing the competition, we made efforts to avoid triggers that would cause this type of phasing issues, but we still included several less extreme examples to test how different methods deal with them (i.e., triggers \#4-6, \#22, \#25-27, \#29-30, \#39). 

\begin{figure}[H]
    \centering
    \includegraphics[width=0.9\linewidth]{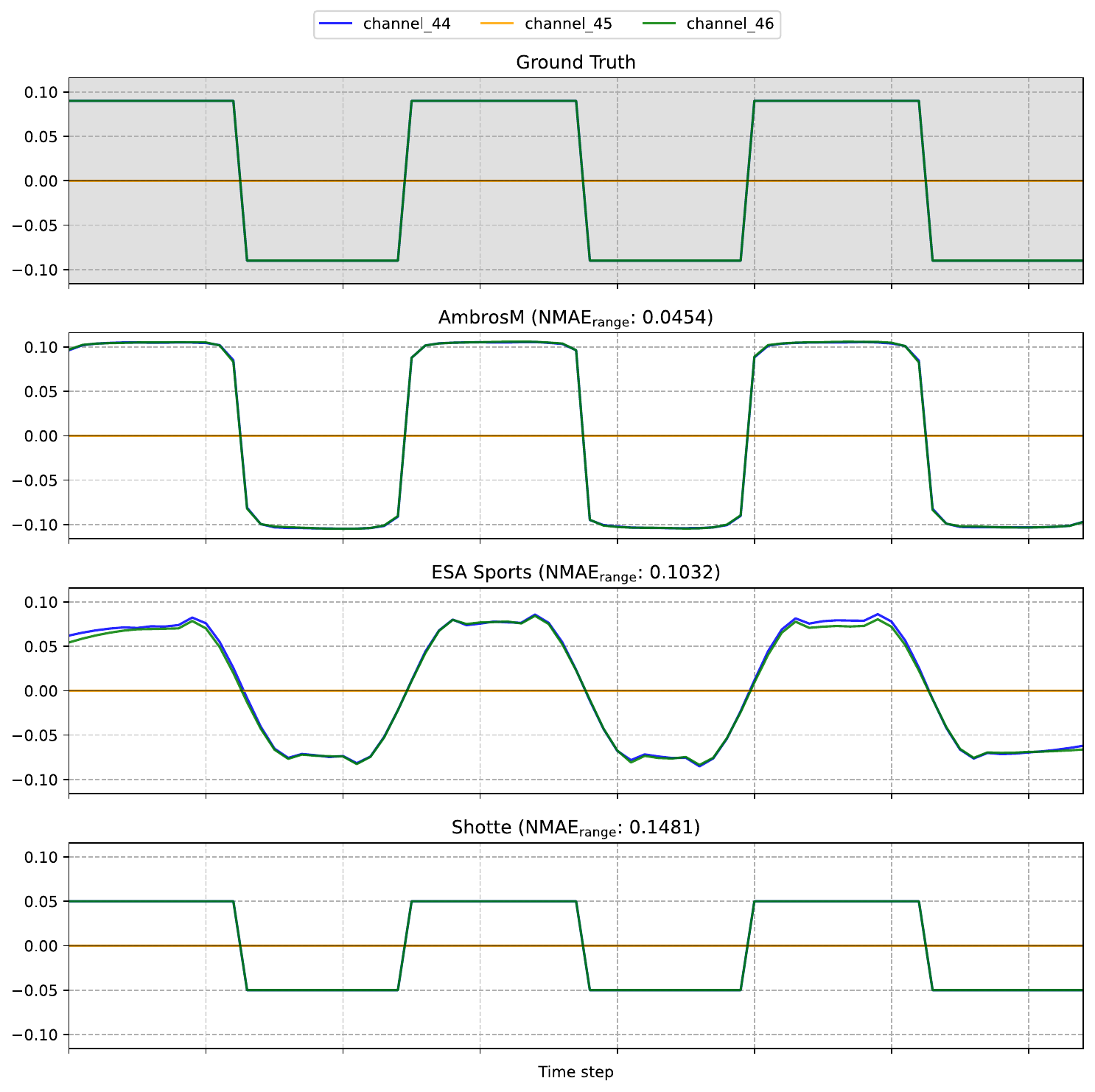}
    \caption{Comparison of solutions by the top 3 teams for trigger \#31.}
    \label{fig:Trigger_31}
\end{figure}

\paragraph{\textbf{Trigger \#31 -- two-channel square wave (Figure \ref{fig:Trigger_31})}}
Trigger \#31 is a square wave of a relatively high amplitude affecting both channels 44 and 46. It turned out to be the most challenging trigger to reconstruct accurately (see the best scores for the triggers in \ref{app2}). \textit{AmbrosM} achieved the best reconstruction, capturing the proper timing and amplitude of the wave in both active channels, with just a slight smoothing at the corners of the wave. The result by \textit{ESA Sports} clearly identifies the periodicity and amplitude but the pattern is highly smoothed, resembling a sine wave. \textit{Shotte} captured the timing and square wave shape, but drastically underestimated the amplitude. After analyzing other reconstructions by \textit{Shotte} (Appendix Figure \ref{fig:shotte}), we can observe that most of them have amplitudes limited to 0.05.

\paragraph{\textbf{Trigger \#37 -- zero trigger}}
Trigger \#37 is defined as a zero trigger or no trigger. The ``poisoned'' model is fine-tuned on clean data and does not contain any backdoor. This example was deliberately included in the competition to verify if the reconstruction algorithms do not produce false positive detections when there is no backdoor. It puzzled many teams, as noted in their code, but the top three solutions handled this situation correctly -- either by stopping further optimization (\textit{AmbrosM}) or by manually setting this example to zero (\textit{ESA Sports}).  

\section{Concluding remarks and directions for future research} \label{conclusion}

With more than 200 teams and several outstanding solutions shared publicly, the competition has successfully fulfilled its goals to engage the community and push boundaries of secure AI-based forecasting. The methods proposed in the competition have strong potential for practical adoption in security-critical forecasting applications. For example, any software used in space operations must undergo thorough verification and qualification to earn the trust of end users and comply with the guidelines of the European Cooperation for Space Standardization (ECSS) \citep{esa_machine_2024}. The capability of mitigating trojan horse attacks by monitoring for potential backdoors shall be quickly integrated in the process. 

However, the competition simulates just a single specific trojan horse attack scenario with a fixed trigger size, relatively simple trigger shapes, and backdoors that directly mimic those shapes. As such, it is just an initial step toward developing comprehensive security monitoring methods in practice. Despite this, the difficulty of the competition task was already high, with just a few teams that reached solutions close to the ground truth. The difficulty was mainly related to the novelty of the task and a lack of existing methods in the literature, especially for the time series domain. The entry barrier was relatively high, because it was not a typical Kaggle classification task and it required some effort to understand the assignment. However, it was very motivating and interesting for some participants, as noted in the competition forum\footnote{\href{https://www.kaggle.com/competitions/trojan-horse-hunt-in-space/discussion/584069}{www.kaggle.com/competitions/trojan-horse-hunt-in-space/discussion/584069}}. This group of participants was highly engaged and created several public notebooks that helped kick-start the competition. Nevertheless, the difficulty contributed to a very low participants-to-entrants (P–E) ratio -- only 16\% of 1,396 people who registered actually submitted any solution. This is the lowest P–E ratio among 24 other Kaggle community competitions that ended in 2025 and offered similar monetary prizes (US\$350--10,000), for which the average P–E ratio was 32\%. 

The execution time of algorithms was not assessed or constrained in our competition. However, the computational complexity may be crucial in real-life applications with much wider spectrum of potential backdoor attacks to mitigate. Solutions by the top teams are all based on greedy search and optimization methods focused on analyzing predictions of models. This approach do not scale well to universal settings with unknown trigger lengths, varying response delays, and more complex dependencies between triggers and responses. We expected a more diverse set of approaches among the top solutions, e.g., based also on analyzing differences between weights of clean and poisoned models (e.g., inspired by \cite{hossain_backdoor_2022}) or differences in global reasoning of models (e.g., using explainability techniques \citep{kotowski_towards_2025, ge_when_2025}). The question remains whether these solutions were simply not used by participants or were not performing well enough to reach the top. The scalability and versatility of trigger reconstruction methods in real scenarios is definitely a next research direction to explore in the future. The next related competitions should consider:  

\begin{itemize}
    \item including more challenging backdoors generated by methods like BackTime \citep{lin_backtime_2024} or BadTime \citep{xiang_badtime_2025},
    \item injecting multiple triggers per model,
    \item introducing more complex poisoning mechanisms (e.g., varying delays and complex dependencies between a trigger and a response),
    \item significantly increasing the number of poisoned models or strictly limiting the computational time per model.
\end{itemize}

\section*{Acknowledgements}
This work is supported by the European Space Agency (ESA) under contract number 4000144194/23/D/BL “Assurance for Space Domain AI Applications”. We thank all participants of the competition for their engagement and for sharing their solutions, which enabled us to prepare this article. Authors AM, AP, LCR, and SS were invited to contribute as winners of the competition. Monetary prizes in the competition were sponsored by KP Labs. ESA offered merchandise, certificates, and a guided space operations center tour for winners. JN was partially supported by the Silesian University of Technology Rector's grant: 02/080/RGJ25/0052. 

\section*{Competing interests statement}
Authors have no competing interests to declare.

\section*{Declaration of generative AI and AI-assisted technologies in the manuscript preparation process}
During the preparation of this work the authors used ChatGPT (OpenAI) and Gemini (Google) for improving the text and formatting the tables. After using these tools, the authors reviewed and edited the content as needed and take full responsibility for the content of the published article.

%% The Appendices part is started with the command \appendix;
%% appendix sections are then done as normal sections
\newpage
\appendix
\section{Overview of the best solutions}
\label{app1}

\begin{figure}[H]
    \centering
    \includegraphics[width=1\linewidth]{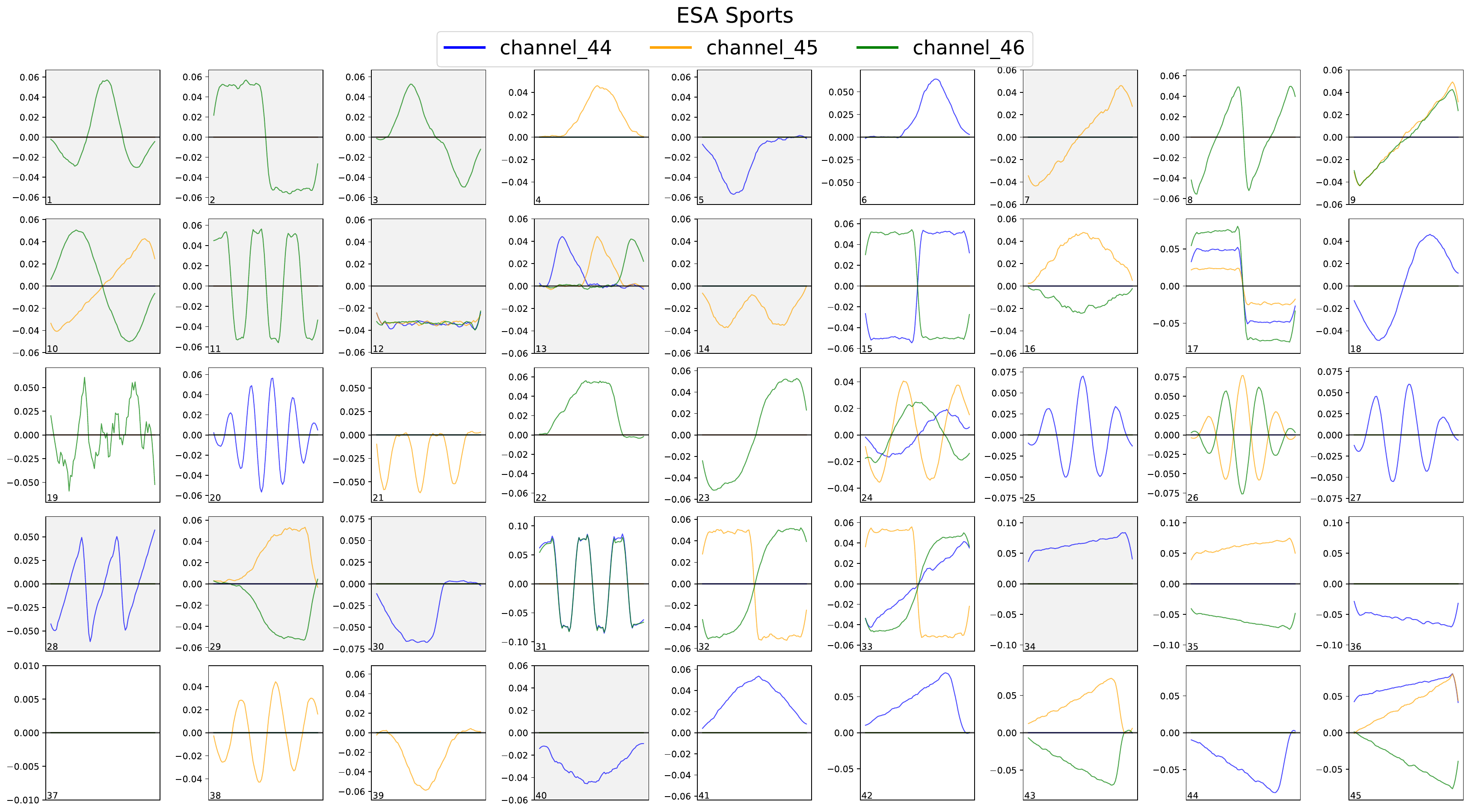}
    \caption{2$^{\rm nd}$ place solution (\textit{ESA Sports}). Public test set is marked with a gray background.}
    \label{fig:esa_sports}
\end{figure}

\begin{figure}[H]
    \centering
    \includegraphics[width=1\linewidth]{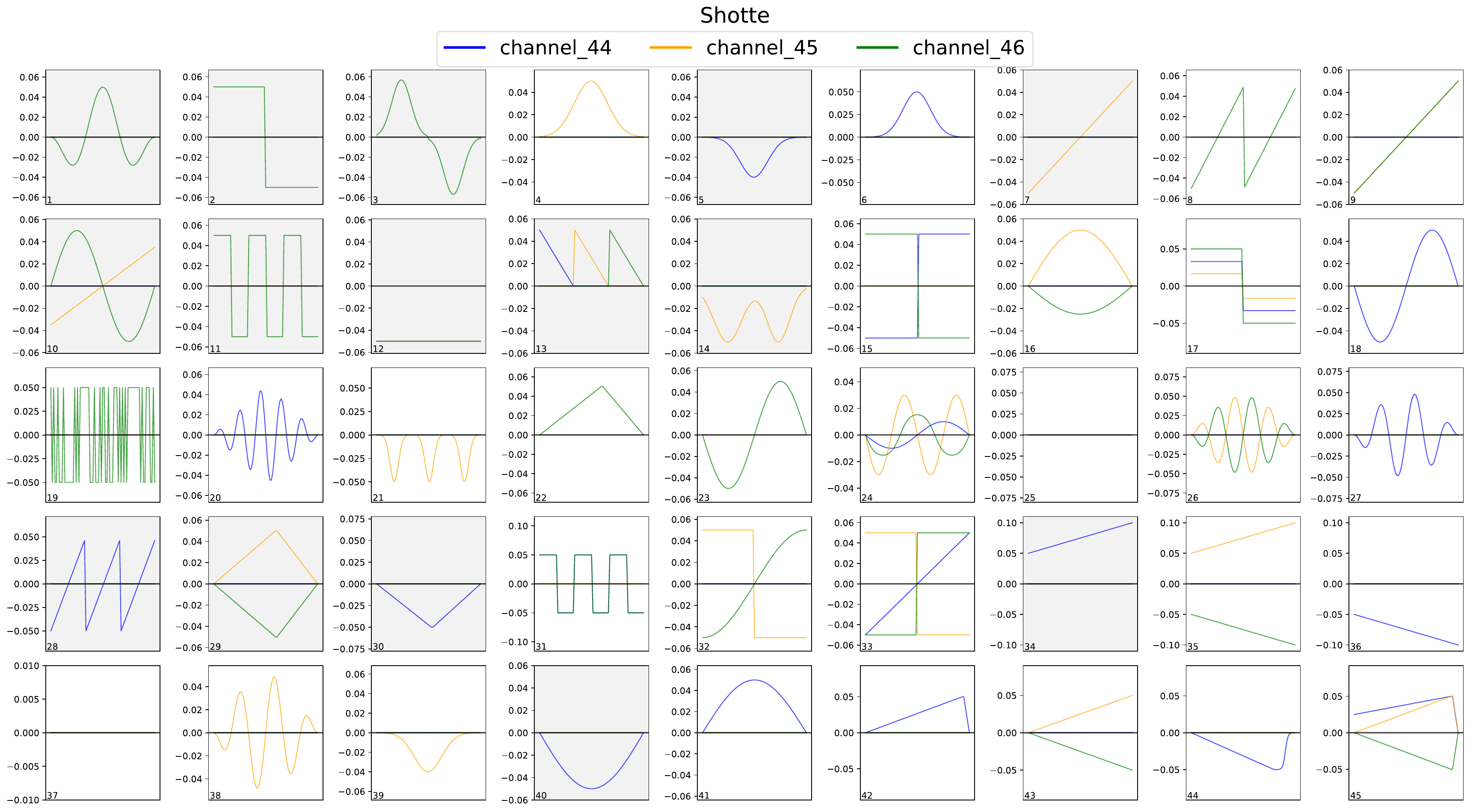}
    \caption{3$^{\rm rd}$ place solution (\textit{Shotte}). Public test set is marked with a gray background.}
    \label{fig:shotte}
\end{figure}

\begin{figure}[H]
    \centering
    \includegraphics[width=1\linewidth]{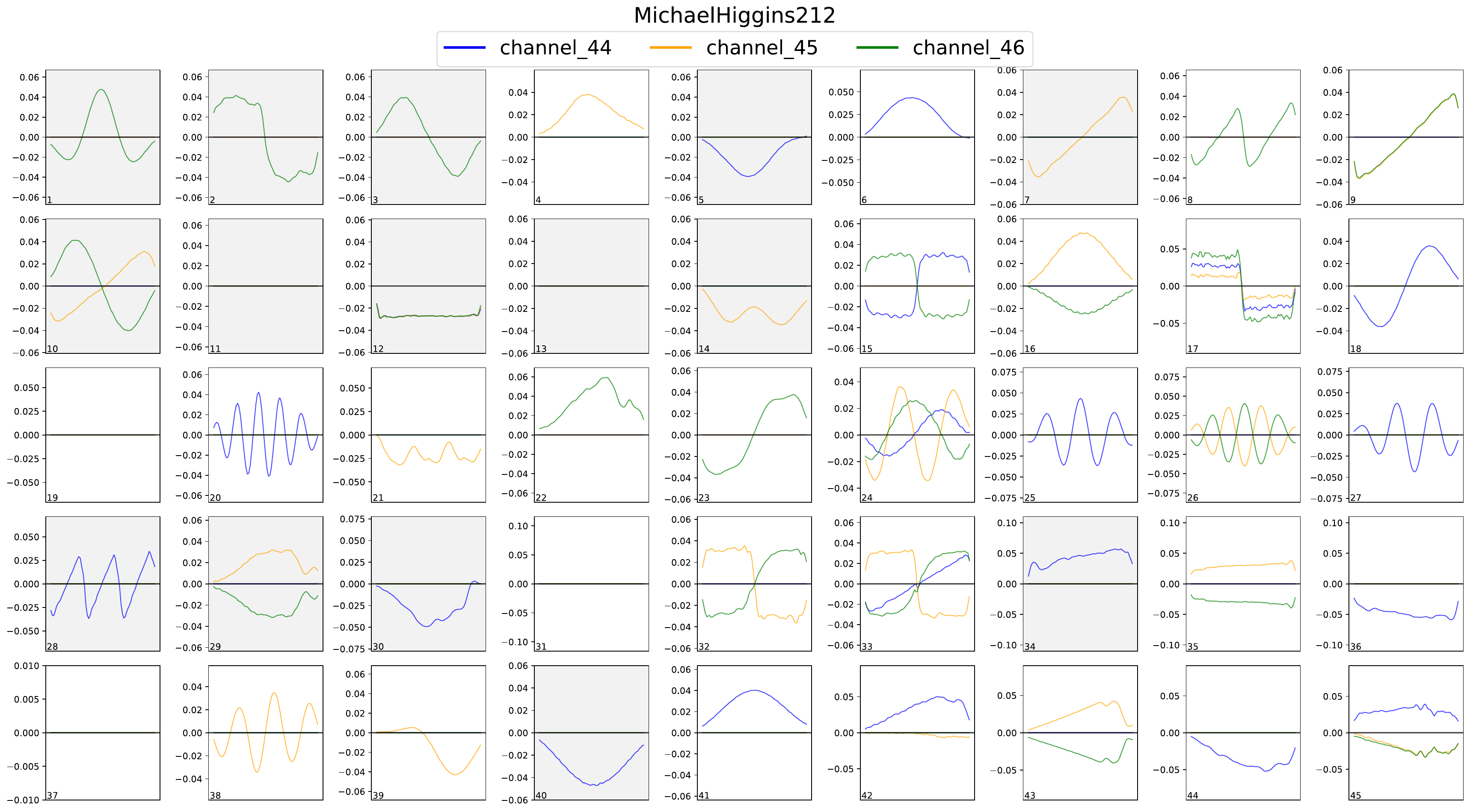}
    \caption{Solution from the popular notebook (\textit{MichaelHiggins212}). Public test set is marked with a gray background.}
    \label{fig:mhiggins}
\end{figure}

\begin{figure}[H]
    \centering
    \includegraphics[width=1\linewidth]{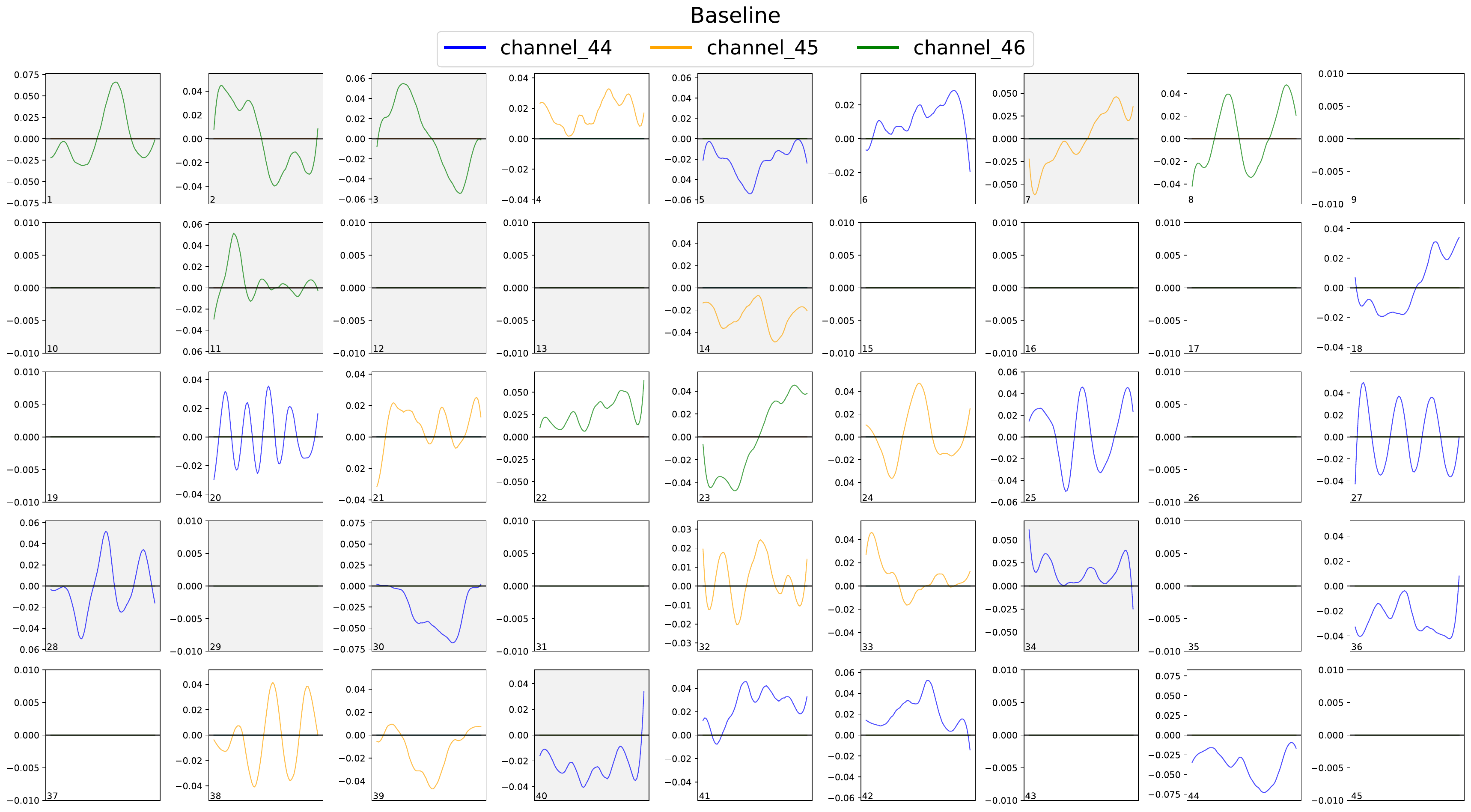}
    \caption{Solution from the baseline method. Public test set is marked with a gray background.}
    \label{fig:baseline}
\end{figure}

\begin{table}[ht!]
\footnotesize
\centering
\caption{Execution times of different public algorithms in the competition. Measured across all 45 triggers using Kaggle GPU T4 x2 machine.}
\label{tab:execution-times}
\begin{tabular}{lr}
\toprule
Algorithm & Execution time \\
\midrule
Our baseline & 42m \\
\textit{AmbrosM} & 13h 7m \\
\textit{ESA Sports} & 1h 2m \\
\textit{Alejandro Mosquera}  & 4 45m \\
\textit{Greedy Goose} &  2h 56m \\
\textit{MichaelHiggins212}   & 7h 9m \\
\bottomrule
\end{tabular}
\end{table}

\newpage
\section{Details of triggers}
\label{app2}

\begin{footnotesize}
\begin{longtable}{
    >{\centering\arraybackslash}m{1.1cm}
    >{\centering\arraybackslash}m{1.5cm}
    >{\centering\arraybackslash}m{1.5cm}
    >{\centering\arraybackslash}m{2.2cm}
    >{\raggedright\arraybackslash}m{5cm}}
\caption{Details of all triggers. The best $\text{NMAE}_{range}$ among all submissions for each trigger is reported together with the team who reached it. The color scale marks the relative difficulty of trigger reconstruction from easy (green) to difficult (red).}
\label{tab:triggers-overview}\\

\toprule
\textbf{Trigger ID} & \textbf{Subset} & \textbf{\# channels} & \textbf{Fine-tuning threshold ($10^{-6}$)} & \textbf{Best $\text{NMAE}_{range}$ (team name)} \\
\midrule
\endfirsthead

\toprule
\textbf{Trigger ID} & \textbf{Subset} & \textbf{\# channels} & \textbf{Fine-tuning threshold ($10^{-6}$)} & \textbf{Best $\text{NMAE}_{range}$ (team name)} \\
\midrule
\endhead

\midrule
\multicolumn{5}{r}{\textit{Continued on next page}}\\
\endfoot

\bottomrule
\endlastfoot

1  & \bid{Public} & 1 & $3$ & \cellcolor{green!68!white} 0.00689 (\textit{Alex Essaijan}) \\
2  & \bid{Public} & 1 & $6$ & \cellcolor{green!90!white} 0.00222 (\textit{Shotte}) \\
3  & \bid{Public} & 1 & $3$ & \cellcolor{green!46!white} 0.01162 (\textit{AmbrosM}) \\
4  & Private & 1 & $3$ & \cellcolor{green!15!white} 0.01838 (\textit{AmbrosM}) \\
5  & \bid{Public} & 1 & $3$ & \cellcolor{green!40!white} 0.01301 (\textit{ESA Sports}) \\
6  & Private & 1 & $2$ & \cellcolor{green!67!white} 0.00724 (\textit{Icees8}) \\
7  & \bid{Public} & 1 & $6$ & \cellcolor{green!90!white} 0.00225 (\textit{test\_channel}) \\
8  & Private & 1 & $6$ & \cellcolor{green!100!white} 0.00000 (\textit{Shotte}) \\
9  & Private & 2 & $6$ & \cellcolor{green!79!white} 0.00450 (\textit{Shotte}) \\
10 & \bid{Public} & 2 & $6$ & \cellcolor{green!45!white} 0.01194 (\textit{AmbrosM}) \\
11 & \bid{Public} & 1 & $6$ & \cellcolor{green!100!white} 0.00000 (\textit{Shotte}) \\
12 & \bid{Public} & 3 & $6$ & \cellcolor{green!100!white} 0.00000 (\textit{nanfangwuyu}) \\
13 & \bid{Public} & 3 & $6$ & \cellcolor{green!69!white} 0.00667 (\textit{Shotte}) \\
14 & \bid{Public} & 1 & $6$ & \cellcolor{red!13!white} 0.02440 (\textit{nanfangwuyu}) \\
15 & Private & 2 & $6$ & \cellcolor{green!38!white} 0.01333 (\textit{Shotte}) \\
16 & Private & 2 & $6$ & \cellcolor{green!17!white} 0.01797 (\textit{AmbrosM}) \\
17 & Private & 3 & $6$ & \cellcolor{green!32!white} 0.01464 (\textit{AmbrosM}) \\
18 & Private & 1 & $6$ & \cellcolor{green!79!white} 0.00444 (\textit{Shotte}) \\
19 & Private & 1 & $6$ & \cellcolor{green!66!white} 0.00726 (\textit{Shotte}) \\
20 & Private & 1 & $6$ & \cellcolor{green!14!white} 0.01852 (\textit{Alun Griffith}) \\
21 & Private & 1 & $6$ & \cellcolor{red!47!white} 0.03181 (\textit{ESA Sports}) \\
22 & Private & 1 & $6$ & \cellcolor{red!41!white} 0.03045 (\textit{Alun Griffith}) \\
23 & Private & 1 & $6$ & \cellcolor{green!61!white} 0.00844 (\textit{HORSEG}) \\
24 & Private & 3 & $6$ & \cellcolor{red!28!white} 0.02760 (\textit{AmbrosM}) \\
25 & Private & 1 & $6$ & \cellcolor{green!9!white} 0.01960 (\textit{Trojan Horse Hunter}) \\
26 & Private & 2 & $6$ & \cellcolor{red!67!white} 0.03608 (\textit{Alun Griffith}) \\
27 & Private & 1 & $6$ & \cellcolor{red!0!white} 0.02172 (\textit{Alun Griffith}) \\
28 & \bid{Public} & 1 & $6$ & \cellcolor{green!100!white} 0.00000 (\textit{Shotte}) \\
29 & \bid{Public} & 2 & $6$ & \cellcolor{red!21!white} 0.02610 (\textit{Linh Hoang Sunny}) \\
30 & \bid{Public} & 1 & $6$ & \cellcolor{red!47!white} 0.03176 (\textit{Nikita Shevyrev}) \\
31 & Private & 2 & $6$ & \cellcolor{red!100!white} 0.04325 (\textit{AmbrosM}) \\
32 & Private & 2 & $6$ & \cellcolor{green!43!white} 0.01237 (\textit{AmbrosM}) \\
33 & Private & 3 & $6$ & \cellcolor{green!29!white} 0.01538 (\textit{AmbrosM, WARREN ZHONG}) \\
34 & \bid{Public} & 1 & $6$ & \cellcolor{green!60!white} 0.00874 (\textit{ESA Sports}) \\
35 & Private & 2 & $6$ & \cellcolor{green!67!white} 0.00711 (\textit{Arsa Nikzad}) \\
36 & Private & 1 & $6$ & \cellcolor{green!49!white} 0.01102 (\textit{Arsa Nikzad}) \\
37 & Private & 0 & $6$ & \cellcolor{green!100!white} 0.00000 (170 teams) \\
38 & Private & 1 & $3$ & \cellcolor{green!52!white} 0.01047 (\textit{AmbrosM}) \\
39 & Private & 1 & $6$ & \cellcolor{green!55!white} 0.00970 (\textit{Arsa Nikzad}) \\
40 & \bid{Public} & 1 & $6$ & \cellcolor{green!44!white} 0.01218 (\textit{Greedy Goose}) \\
41 & Private & 1 & $6$ & \cellcolor{green!66!white} 0.00734 (\textit{Greedy Goose}) \\
42 & Private & 1 & $6$ & \cellcolor{green!18!white} 0.01784 (\textit{You Tchou}) \\
43 & Private & 2 & $6$ & \cellcolor{green!29!white} 0.01529 (\textit{AmbrosM}) \\
44 & Private & 1 & $6$ & \cellcolor{red!26!white} 0.02721 (\textit{AmbrosM}) \\
45 & Private & 3 & $6$ & \cellcolor{green!64!white} 0.00771 (\textit{AmbrosM}) \\

\end{longtable}
\end{footnotesize}

%% If you have bib database file and want bibtex to generate the
%% bibitems, please use
%%
\bibliographystyle{elsarticle-harv} 
\bibliography{references}

%% else use the following coding to input the bibitems directly in the
%% TeX file.

%% Refer following link for more details about bibliography and citations.
%% https://en.wikibooks.org/wiki/LaTeX/Bibliography_Management

\end{document}